\documentclass[journal,twoside]{IEEEtran}
\usepackage{cite}
\usepackage{amsmath,amssymb,amsfonts}
\usepackage{algorithmicx}
\usepackage{graphicx}
\usepackage{textcomp}

\usepackage{amsmath}
\usepackage{amsfonts}
\usepackage{algorithm}
\usepackage{algpseudocode}
\usepackage{array}
\usepackage{textcomp}
\usepackage{stfloats}
\usepackage{url}
\usepackage{verbatim}
\usepackage{graphicx}
\usepackage{amssymb}
\usepackage{amsthm}
\usepackage{comment}

\graphicspath{figs}
\usepackage{colortbl}
\usepackage{enumitem}
\usepackage{booktabs}
\usepackage{hyperref}
\usepackage{tabularx}

\usepackage{multirow}
\usepackage{xcolor}
\usepackage{ulem}

\newcolumntype{C}{>{\centering\arraybackslash}X}

\usepackage[numbers]{natbib}
\usepackage{cleveref}
\crefname{figure}{Fig.}{Fig.}
\crefname{table}{Table}{Tables}
\crefname{algorithm}{Algorithm}{Algorithms}
\crefname{section}{Section}{Sections}
\crefname{equation}{}{}

\usepackage{float}
\def\BibTeX{{\rm B\kern-.05em{\sc i\kern-.025em b}\kern-.08em
    T\kern-.1667em\lower.7ex\hbox{E}\kern-.125emX}}
\begin{document}

\title{Reinforcement Learning Based Bidding Framework with High-dimensional Bids in Power Markets}

\author{Jinyu Liu, \IEEEmembership{Student Member, IEEE}, Hongye Guo, \IEEEmembership{Member, IEEE}, Yun Li, Qinghu Tang, \IEEEmembership{Student Member, IEEE}, Fuquan Huang, Tunan Chen, Haiwang Zhong, \IEEEmembership{Senior Member, IEEE} and Qixin Chen, \IEEEmembership{Senior Member, IEEE}

\thanks{This work was supported in part by the Science and Technology Project of China Southern Power Grid under Grant 090008KC23020006 }
\thanks{
Jinyu Liu, Hongye Guo (Corresponding Author, e-mail: hyguo@mail.tsinghua.edu.cn), Qinghu Tang, Haiwang Zhong, and Qixin Chen are with the State Key Laboratory of Power Systems, Department of Electrical Engineering, Tsinghua University, Beijing 100084, China. 

Yun Li, Fuquan Huang, and Tunan Chen are with the Shenzhen Power Supply Co. Ltd., Shenzhen 518001, China.
Tunan Chen is also with the Tsinghua Shenzhen International Graduate School, Shenzhen 518071, China.
}
}
\maketitle

\begin{abstract}

Over the past decade, bidding in power markets has attracted widespread attention.
Reinforcement Learning (RL) has been widely used for power market bidding as a powerful AI tool to make decisions under real-world uncertainties.
However, current RL methods mostly employ low dimensional bids, which significantly diverge from the \textit{N price-power pairs} commonly used in the current power markets. The N-pair bidding format is denoted as High Dimensional Bids (HDBs), which has not been fully integrated into the existing RL-based bidding methods.
The loss of flexibility in current RL bidding methods could greatly limit the bidding profits and make it difficult to tackle the rising uncertainties brought by renewable energy generations.
In this paper, we intend to propose a framework to fully utilize HDBs for RL-based bidding methods. 
First, we employ a special type of neural network called Neural Network Supply Functions (NNSFs) to generate HDBs in the form of N price-power pairs.
Second, we embed the NNSF into a Markov Decision Process (MDP) to make it compatible with most existing RL methods.
Finally, experiments on Energy Storage Systems (ESSs) in the PJM Real-Time (RT) power market show that the proposed bidding method with HDBs can significantly improve bidding flexibility, thereby improving the profit of the state-of-the-art RL bidding methods.

\end{abstract}


\section{Introduction}
\label{sec:introduction}

\IEEEPARstart{M}{arket} uncertainty is becoming a crucial aspect of power market bidding.\cite{kempitiya_artificial_2020}
In recent years, the variability in power generation has led to more fluctuating day-ahead (DA) and real-time (RT) market prices\cite{cevik_chasing_2023}. The increase in market price uncertainty poses a greater challenge for market bidding.

This trend of rising uncertainty is expected to persist in the following decades due to the increase in renewable generations\cite{rai_impact_2020}. The International Renewable Energy Agency (IRENA) has projected that the worldwide installed capacity of renewable energy sources will increase significantly in the coming years.
Currently, renewables account for approximately 30\% of the total generation capacity. This figure is projected to surpass 50\% by 2035\cite{international_renewable_energy_agency_renewable_2023}. The future power system will be more reliant on renewables, and the power markets will face higher uncertainties.

To achieve high profits under high market uncertainties, generation companies (GENCOs) must wisely strategize their market bids to ensure market profits. 
The topic of strategic bidding has been a hotspot of research for the last decade. To tackle the rising uncertainties in power markets, RL-based methods have been widely adopted to tackle the real-world market uncertainties.

RL is a subset of Machine Learning (ML), focusing on training algorithms to make a sequence of decisions by interacting with an environment to achieve a goal, where success is measured by a system of rewards and penalties. 

% RL-based methods are able to tackle real-world market uncertainty by learning the parameters of the decision policies based on historical market data.
% In detail, RL methods, such as deep Q-learning\cite{song_prioritized_2021}, policy gradient\cite{ye_deep_2020}, Actor-Critic\cite{huang_deep-reinforcement-learning-based_2021}, and Supervised AC\cite{badoual_learning-based_2021}, manage price uncertainties by directly learning a bidding strategy in simulated real-world bidding processes. 

Past RL-based bidding methods have been applied to multiple-market\cite{bian_optimal_2024} and multiple-subject\cite{wei_self-dispatch_2022} bidding scenarios, both under real-world uncertainties. Although RL-based bidding methods can address real-world uncertainties, they have a fundamental drawback: they cannot effectively utilize HDBs in bidding.

\relax{
The HDB is the most common bid format in real-world power markets.
An HDB consists of several price-power pairs arranged in monotonic increasing order\cite{PJM-Energy-and-Ancillary-Service-Market-Operations}.
GENCOs use these bidding pairs to indicate how much power they are willing to commit to a certain market price, especially when the price is highly volatile.
In the PJM\cite{PJM-Energy-and-Ancillary-Service-Market-Operations}, CAISO\cite{caiso_real-time_2023}and AEMO\cite{australian_energy_market_operator_integrating_2022} power market, N equals 10, which means 10 dimensions are used to represent bidding prices, and another 10 dimensions are used to represent bidding quantities.
Because these bidding pairs are formulated in a high-dimensional space (2$N$ is usually 20), we refer to this real-world bid format as the High-Dimensional Bids (HDBs) in this paper, compared with the Low-Dimensional Bids (LDBs) used in most current studies.
}

\relax{
The HDBs in bidding strategies are important for GENCOs to cope with the rising price uncertainty. }As prices are hard to predict very accurately in certain real-world markets\cite{zhang_predicting_2022,yang_qcae_2022}, it is difficult for GENCOs to schedule the generation plan beforehand, especially for new-type market participants such as Energy Storage System (ESS) and Virtual Power Plant (VPP).
\relax{
In this case, GENCOs can use HDBs to indicate their generation plan for N in different market price ranges. So that they can achieve a satisfactory market clearing result at every market clearing price.
The HDB is a useful tool to express GENCOs' willingness on the market prices in the market clearing process.
}

Currently, HDBs (High-dimensional Bids) are mostly considered by optimization-based approaches, but rarely utilized in RL-based methods.\color{black}
Optimization-based methods are able to utilize HDB's high dimensionality by optimizing the bid parameters by modeling the possible market outcomes.
They mostly utilize stochastic programming\cite{wang_two-stage_2024,herding_stochastic_2023}%wang_day-ahead_2023
and robust programming\cite{najafi-ghalelou_maximizing_2024,karasavvidis_optimal_2024} to optimize the bid parameters.
Because optimization-based methods can fully utilize the known models for bidding,  they are outstanding for market scenarios that can be completely modeled.
% \cite{herding_stochastic_2023} proposed a two-stage stochastic programming approach to optimize the HDBs in the ERCOT day-ahead market.
However, many optimization-based methods\cite{wang_two-stage_2024,herding_stochastic_2023,najafi-ghalelou_maximizing_2024} rely on price predictions to schedule the generator in the rea world, so the precision of price predictions will obviously affect the effectiveness of the bidding strategies.
The performance of optimization-based methods bidding has been surpassed by RL-based methods in high-uncertainty scenarios \cite{jeong_deep_2023,li_temporal-aware_2024}, including the real-time market ESS bidding studied in this paper \cite{li_temporal-aware_2024}.
Because the price forecast accuracy is low in such a case\cite{zhang_predicting_2022,yang_qcae_2022}, it is difficult for optimization-based methods to bid effectively.
\color{black}

RL-based methods can tackle real-world uncertainties by learning real-world market data, but they currently can only utilize Low-Dimensional Bids (LDBs), like one-value price bids\cite{li_temporal-aware_2024,wei_self-dispatch_2022,anwar_proximal_2022}, one-value power bids\cite{esmaeili_aliabadi_emerging_2022,wang_adaptive_2024,du_approximating_2021}, one-pair price-power bids\cite{jeong_deep_2023,tao_deep_2022,ren_reinforcement_2023} etc. These LDBs are all greatly simplified versions of HDBs.
For example,
\cite{li_temporal-aware_2024}-\cite{anwar_proximal_2022} use a bidding format of one price bid with fixed power, and the clearing result is either zero power or full power.
\cite{esmaeili_aliabadi_emerging_2022}-\cite{du_approximating_2021} use one-power bids, which means they decide their power output regardless of actual market prices. 
\cite{jeong_deep_2023}-\cite{ren_reinforcement_2023} use the bid format of one price-power pair, where the GENCO submits a price threshold for generating power, and a power quantity to indicate how much power it is willing to generate. 
Such LDBs are greatly simplified from the HDBs in real-world power markets. The loss of flexibility in market bids can cause loss in the bidding performance\cite{herding_stochastic_2023,wang_two-stage_2024}.

Though these bidding formats are easier to learn using RL-based methods, they sacrifice the expressiveness of the market bids. They only have 0$\sim$2 price/power bids, and cannot fully reflect the GENCOs' willingness on the market prices.

There have been a few RL-based bidding methods that have attempted to adopt the HDB format\cite{jia_reinforcement-learning-based_2022,liang_agent-based_2020,pedasingu_bidding_2020,wang_agent_2019}, but usually in an indirect way. In detail, they fail to generate HDBs in the original high-dimensional and continuous HDB space, which ignores the most important features of HDB format in bidding.
Instead, these methods use case-specific simplifications to transform the high-dimensional bidding space into low-dimensional spaces, which inevitably hurts the degree of freedom of bids.
For example, 
 \cite{jia_reinforcement-learning-based_2022,mallaki_strategic_2021} use the overflow proportion as a low-dimensional continuous bidding space, where they decide the overflow ratio of the actual thermal generator cost curve.
 \cite{liang_agent-based_2020} assumes that the HDBs are in the space of an affine function and decides the slope of this affine bidding function.  
 \cite{pedasingu_bidding_2020} make decisions in a discrete space consisting of nine HDB samples and convert the bidding problem to a nine-option decision problem.% subramanian_learn_2019
 \cite{wang_agent_2019}  designs a custom set of decision variables and converts the bidding space to a 3-dimensional continuous space. 

These methods utilize HDBs in an indirect way and have the following defects:
First, they cannot make decisions in the original HDB space and cannot fully utilize the potential of HDBs.
Second, they require specific parameterization designs for specific bidding problems, which may hinder their ability to adapt to new entities and market conditions. 
Third,  because the parameterization designs are manually specified, they could be suboptimal and limit the bidding performances.

To overcome the above-mentioned challenges, this paper intends to propose an HDB generation framework for RL-based methods. The proposed framework can serve as a tool for various types of bidding entities to better utilize HDBs for market bidding in real world.
However, several challenges are involved in achieving the above aims.
Firstly, we need to generate HDBs that are suitable for RL-based bidding methods. This includes encoding the bids in a format that captures the essential information needed for decision-making in the bidding process while ensuring that the representation is compact enough to facilitate quick learning and actioning by the RL model.
Secondly, we need to ensure that the generated HDB satisfies market bidding requirements, such as bidding pairs' monotonicity of HDBs. This ensures that the final bidding strategy is not only optimized for profitability or cost-effectiveness but also complies with market rules and expectations.
Thirdly, the proposed framework should be adaptable to different fuel types and RL methods so that it can be tailored to meet a wide range of requirements and scenarios in the energy market bidding landscape.
To the best of our knowledge, there has not been an RL-based bidding algorithm that can generate the HDBs without simplification. 

In this paper, we propose an HDB generation framework that is compatible with most RL algorithms. It supports efficient and effective HDB bidding under high uncertainties.
First, we identify a special type of neural network with price input and power output, which is called the Neural Network Supply Function (NNSF).
Second, we extract HDBs from the NNSFs' input-output relationship. The extracted HDBs will satisfy the market bidding requirements, and will keep the essential bidding strategy of the NNSF.
Third, we approximate the HDB bidding process with NNSF and use the approximation to propose a training process that is suitable for most RL training frameworks.
Finally, we conduct experiments on Energy Storage Systems (ESSs) in real-world power market, such as PJM. We demonstrate that the proposed HDB generation method can improve the bidding performance of the state-of-the-art RL-based bidding methods.

The main contributions of this paper are summarized as follows:

\begin{itemize}
    \item A neural network based modeling method for HDB bidding has been proposed, which exhibits three key characteristics: it achieves lossless representation of HDBs, meets the feature requirements of real market bids, and facilitates integration into RL for training.

    \item This paper proposes a framework that combines HDB with RL methods. It includes the training and testing phases. This framework is compatible with the majority of RL methods and facilitates improvements in HDB biddings.
    
    \item  The proposed method's effectiveness is shown in Energy Storage Systems (ESS) in the real-world PJM RT power market. The proposed HDB bidding algorithm can significantly improve bidding flexibility and improve bidding arbitrage performance by 15.40\% compared with the RL-based bidding methods that utilize LDBs.

\end{itemize}

The rest of this paper is organized as follows. In \cref{sec:problem_formulation}, we describe the bidding problems. In \cref{sec:HDBgen}, we generate HDBs from the input-output relationship of NNSFs. In \cref{sec:HDBlearn}, we learn NNSF with RL algorithms. In \cref{sec:experiments}, we conduct performance evaluations. Finally, conclusions are made in \cref{sec:conclusion}.

\section{System Model and Problem Formulation}
\label{sec:problem_formulation}

In this paper,  we consider an ESS that uses HDBs to participate in the real-world RT energy market.
First, we will describe the HDB clearing model and the ESS model, then formulate the HDB bidding problem of the RT energy market.

\subsection{HDB based Market Bidding Model}

An HDB consists of a series of $N$ price-power pairs that increase in a monotonically ascending manner. 
We denote HDBs with  the price bids, denoted as $\bar{\lambda}_{i}$, and the power bids, denoted as $\bar{p}_{i}$. 
The HDB needs to satisfy the following constraints:

\begin{equation}\label{energy_bid_rule}
\begin{aligned}\lambda _{\min} \leqslant \bar{\lambda} _{i} \leqslant \lambda _{\max} & \\p_{\min} \leqslant \bar{p}_{i} \leqslant \ p_{\max} & ,\ \text{for} \ i=1,\dotsc ,N\\\bar{\lambda} _{i} \leqslant \bar{\lambda} _{i+1} & \\\bar{p}_{i} \leqslant \bar{p}_{i+1} & ,\ \text{for} \ i=1,\dotsc ,N\end{aligned}
\end{equation}

A market bidding pair $(\bar\lambda_i,\bar p_i)$ is accepted by the power market if its bidding price is lower than the market clearing price.
The power of the largest accepted bidding pair will be the market clearing power of the bidder.

In power markets, one HDB could be cleared for multiple rounds. For example, in the RT energy market, an HDB is submitted hourly, and the market clearing is conducted for the next 12$\times$5 minutes. Therefore, all HDB values can affect the clearing results of a period.

\subsection{ESS Model}
In this subsection, we propose the ESS model based on lithium-ion batteries.
With the significant decrease in battery production costs, batteries can now economically power major energy shifts.
The ESS participates in the RT energy market, which is one of the most profitable energy markets for energy arbitrage. The proposed ESS model considers its state transition and its bidding objective.

The internal state of an ESS is changed according to the State of Charge (SoC) model . 
It can be described in time-step formats:

\begin{equation}\label{ESSmodel}
\begin{aligned}
SoC_{t} = & SoC_{t-1} +\tau (\eta ^{c} p^c_{t} -\frac{p^d_{t}}{\eta ^{d}} )\\
0\leqslant  & SoC_{t} \leqslant SoC_{\max} ,\\
0\leqslant  & p^c_{t}  \leqslant p_{\max} ,\\
0\leqslant  & p^d_{t}  \leqslant p_{\max} ,\\
 & p^c_{t} \cdot p^d_{t} =0
\end{aligned}
\end{equation}

where $\tau$ is the timestep, usually 5 minutes for the RT market. $\eta^c, \eta^d$ represents the charging and discharging efficiency of the ESS, and $p^c_{t}$, $p^d_{t}$ are the charging and discharging power.
$p^c_{t}$ and $p^d_{t}$ cannot be non-zero at the same time. 

\color{black}%was black
The bidding objective of an ESS consists of three parts, including the market bidding income, the ESS depreciation cost, and the SoC penalty rewards: 

\begin{equation}
\label{equ:ESS_reward}
    r_t = \Sigma_t \lambda_t \cdot p^d_t  +\Sigma_t r^{dep}_t + \Sigma_tr_t^{ {soc\_violate }}
\end{equation}

The first term is the market bidding income based on clearing results of \cref{energy_bid_rule}, where $\lambda_t$ is the energy market clearing price.
The second term is the  ESS's depreciation costs. Since the modeling of ESS is not the focus of this paper, we consider a typical degradation cost that is proportional to the discharge power and time elapsed\cite{xu_operational_2020}. The cost model can be expressed formally as:

\begin{equation}
r^{dep}_t=-\lambda^{dep}\cdot p^d_t\cdot \tau
\end{equation}
where $\lambda^{dep}$ is the depreciation cost constant. The negative sign means the reward for depreciation is always negative. 

The third term is the SoC violation penalty of the ESS. In the RL training process, the RL actor's actions could lead to an invalid SoC. In such cases, an additional penalty is added to the RL training process to avoid SoC violations. The SoC violation penalty can be expressed as:

\begin{equation}
r_t^{ {soc\_violate }}= 
\begin{cases}
0 & ,0 \leq SoC_{t+1} \leq SoC_{max} \\
-P & ,\text { else }
\end{cases}
\end{equation}
where $P$ is the reward penalty constant.

\color{black}

\subsection{Bidding Problem Formulation}
\label{sec:bid_problem_formulation}

% The main challenge of ESS bidding is its non-generating characteristic and the limited storage capacity. 
% First, ESS's explicit cost of generating power closely relates to its charging history. Because ESS cannot generate net power, they must charge power before discharge. Therefore, the explicit cost of ESS discharge depends on its charging strategy and changes with the market prices.
% Second, ESSs need to make bidding decisions considering multiple future timesteps, usually in the horizon of multiple days.
% Because the ESS has limited energy storage capacity, their power actions are temporally correlated. Therefore, each action will have an opportunity cost due to its influence on the ESS's ability to commit power actions in the upcoming days. 
% Because the ESS opportunity cost depends on future market prices, the opportunity cost is hard to estimate when future prices are highly uncertain.
% In all, ESS needs to bid strategically to reach a high profit under uncertain market prices.

Next, to solve the bidding problem with RL, we will describe the bidding problems of the ESS using Markov Decision Processes (MDPs).
The MDP is a mathematical framework used to model decision processes where outcomes are influenced by both randomness and the decisions made by the decision-maker. An MDP is defined by a tuple $(\mathcal{S},\mathcal{A},\mathcal{R},\mathcal{T},\gamma)$, consisting of state, action, reward, transitions, and discount factor. \color{black}
At each time step, the agent observes the current state of the environment $s_t\in \mathcal{S}$ and takes an action $a_t\in\mathcal{A}$. The environment transitions to a new state $s_{t+1}$ by transition $\mathcal{T}$, and the agent receives a reward signal $r_t(s_t,a_t)\in\mathcal{R}$ that indicates the desirability of the state-action pair. The goal of the agent is to learn a policy $\pi$, which is a mapping from state to actions, that maximizes the cumulative reward $\Sigma_{t=1}^\infty \gamma^tr_t(s_t,a_t)$ over time. The discount factor $\gamma$ is between $0$ and $1$, which measures the importance of futrue rewards and instant rewards.\color{black}
The ESS bidding MDP is formulated as follows. It has a state $\hat s_t$, which consists of market observation and energy levels by the bid submission time. (We use $\hat s_t$ instead of $s_t$ because the state will be augmented later) The action $a_t$ is the HDB \cref{energy_bid_rule}, which is the output of the bidding policy. The reward $r_t$ is the bidding objective \cref{equ:ESS_reward}. The transition probability $\mathcal{T}$ consists of two parts. The first part is the ESS's energy level transitions (2). The second part is the market price transition, \color{black}which is influenced by multiple market factors. Because the second part of the transition probability is complicated and hard to model, we will not formulate it explicitly. Instead, we use real-world data of this transition probability, which are the real-world market price histories to simulate the transition. A certain portion of historical prices are used to simulate the training set, and a certain portion of more recent prices are used to simulate the testing set.
\color{black}

The bidding objective is to maximize the total reward \cref{equ:ESS_reward} with the HDBs \cref{energy_bid_rule}.

\section{Neural Network based HDB Generation}
\label{sec:HDBgen}

In this section, we propose an HDB generation framework that generates HDB actions from market states by employing a special type of neural network called NNSF.
 The generated HDBs will satisfy the HDB bidding rules \cref{energy_bid_rule} and can be submitted for market bidding.
The HDB generation framework further enables neural network training with RL of \cref{sec:HDBlearn}.

\begin{figure}[h]
    \centering
    \includegraphics[width=1\linewidth]{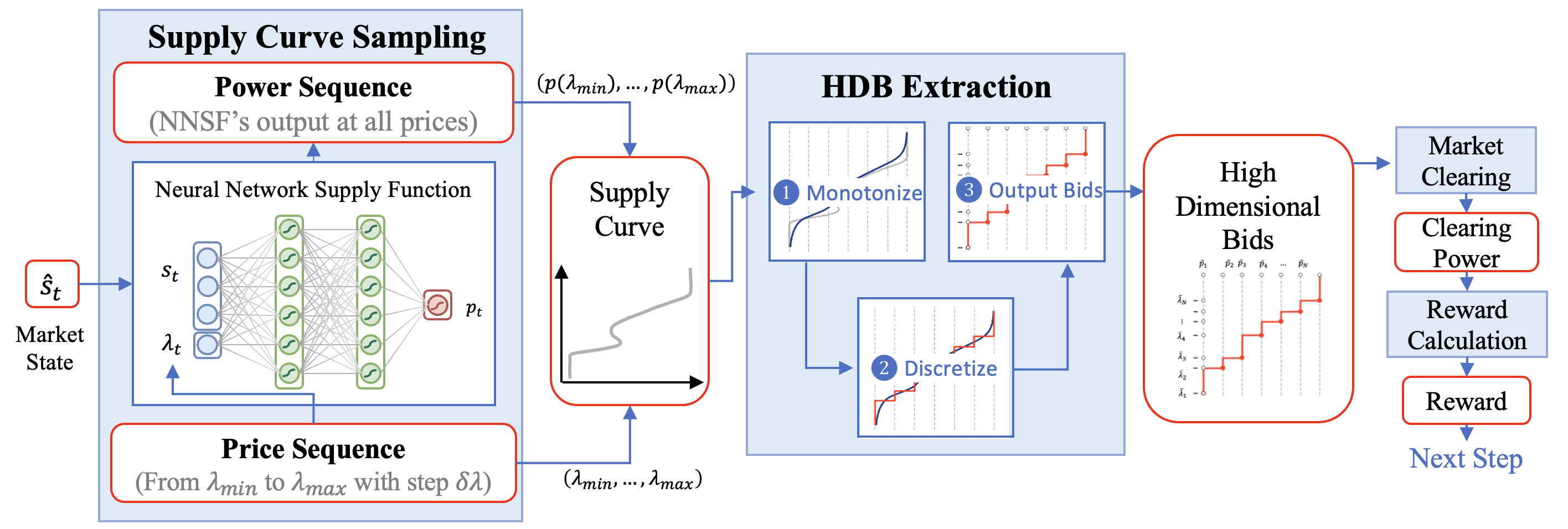}
    \caption{The proposed HDB generation framework}
    \label{fig:generation_framework}
\end{figure}

The proposed framework to generate HDBs from neural networks is shown in \cref{fig:generation_framework}. First, a supply curve is sampled from the NNSF's input-output relationship in the \textit{Supply Curve Sampling} process. In this process, the NNSF's output is sampled multiple times with different input prices (from $\lambda_{\min}$ to $\lambda_{\max}$ with step size $\delta \lambda$). The corresponding price-power pairs constitute the supply curve. Second, a $2N$ dimensional HDB is extracted from the supply curve in the \textit{HDB Extraction process}. Three steps are introduced to extract an HDB from the supply curve, which are the monotonize, discretize, and output steps. The extracted HDB will satisfy the bidding rule \cref{energy_bid_rule}. Third, the HDB is submitted to market clearing, and the market operator computes the clearing price and clearing power result and returns the bidding reward. Finally, we proceed to the next bidding step.

\color{black}
The main idea of the proposed HDB generation framework can be understood as follows.
An infinite dimensional HDB (when $N\rightarrow\infty$) is a continuous supply curve, which continuously maps price to power.
We use a continuous function (the NNSF) to represent the supply curve so that it can be improved by neural network training (\cref{sec:HDBlearn}).
Because the 2N-dimensional HDBs are downsampled from infinite-dimensional supply curves, the 2N-dimension flexibility can be utilized.
Past methods can achieve bidding flexibility of a maximum 4-dimension\cite{tao_deep_2022}, and the proposed method can achieve 2N-dimension so that the bidding flexibility is significantly improved.
\color{black}

In this section, we will first introduce the NNSF in \cref{sec:NNSF}. Then, we propose the supply curve sampling process in \cref{sec:supply_curve_sampling} and the HDB extraction process in \cref{sec:bid-generation}.  At last, the suitability of the proposed HDB generation framework to RL algorithms is discussed in \cref{sec:RL-suitability}.

\subsection{Neural Network Supply Function}\label{sec:NNSF}

In this paper, we use a special class of neural networks to generate supply curves, which we name the NNSF.

In economics, the supply curve (or supply function) is a fundamental concept that represents the relationship between the quantity of a good or service that producers are willing and able to sell and its price.
% In market bidding, HDB serves as a supply curve. As shown in \cref{fig:supply_function_explain}, HDBs are market bids in the form of $N$ \textit{price-power pairs}, and the $N$ pairs can be interpreted as a step-wise supply curve. For each market \textit{price}, we can get the producers' \textit{power} quantity.

In machine learning, neural networks can be used to represent any arbitrary function and neural networks are used as versatile function approximators. Therefore, they can be used to represent the supply curves (the supply function) of a bidder.

\begin{figure}[h]
    \centering
    \includegraphics[width=0.8\linewidth]{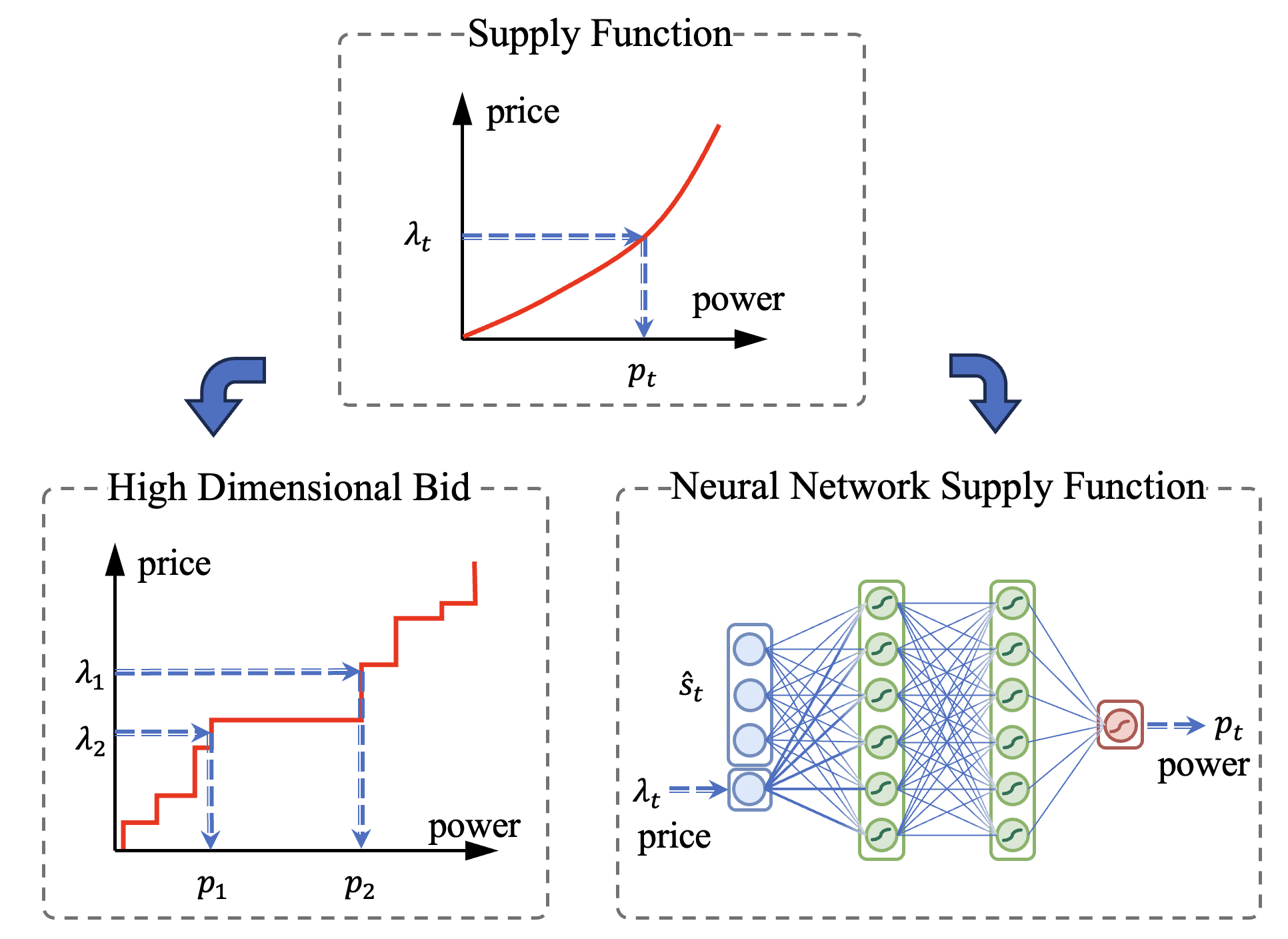}
    \caption{HDB and NNSF are both Supply Functions}
    \label{fig:supply_function_explain}
\end{figure}

Using these observations, we define neural networks that have price inputs and power outputs as NNSFs. By definition, an NNSF maps price to power, and the mapping is a neural network. 
\color{black}Additionally, the shape of the supply curve is also determined by various factors, such as the ESS state, the time of bidding, the market price trend, etc. Therefore, \color{black}
NNSF has an additional input of market state $\hat{s}_t$.

In the following context, we will denote the NNSF as $p_t = \pi_\theta(\hat{s}_t,\lambda_t)$, where $\pi$ is the neural network, $\theta$ is the neural network parameter, $\hat{s_t}$ is the market state input, and $\lambda_t$ is the market price input. $\hat{s_t}$ and $\lambda_t$ consist of the full input (full state) $s_t$ of the NNSF. The output of the NNSF is the output power $p_t$.

\subsection{Supply Curve Sampling} \label{sec:supply_curve_sampling}

The first step of generating HDBs is sampling a supply curve from the NNSF.  The supply curve ($\lambda-p$) is the NNSF's input-output relationship at a certain market state input $\hat{s}_t$.  The supply curve is a point set consisting of $M$ price-power pairs. The schematic of the supply curve sampling process is shown in \cref{fig:generation_framework}.

\color{black}
The supply curve is derived by sampling the neural network's output value across the whole price range. \color{black}
The price range $[\lambda_{\min},\lambda_{\max}]$ is divided into $M$ segments with step size $\delta \lambda$. The NNSF's power output at each segment is computed: $p_{(\cdot)} = \pi_\theta(\hat{s}_t,\lambda_{(\cdot)})$. Finally, the NNSF's price-power relation is sampled as the supply curve: $[(\lambda_{\min},...,\lambda_{\max}),(p(\lambda_{\min}),...,p(\lambda_{\max}))]$.

\color{black}
Note that the supply curve sampling process can be computed in parallel, avoiding $\mathcal{O}(M)$ time complexity. Since neural networks are typically deployed on GPUs designed for parallel computations, the network sampling operation can use a batch size of $M$. If the GPU computes the entire batch in parallel, the computation time complexity is $\mathcal{O}(1)$. In our experiments, we will demonstrate that this process is computationally efficient.\color{black}

\subsection{HDB Extraction}
\label{sec:bid-generation}

In this subsection, we will extract HDBs from the supply curves.

The supply curve and the HDB are both supply functions. However, HDBs are supply functions that have special requirements.
To be specific, HDBs are \textit{monotonic} and \textit{discrete-output} supply functions. The HDB  bidding rule \cref{energy_bid_rule} defines a special family of supply functions. First, the supply function should be \textit{monotonic}, so that the bidding power increases with the market price. Also, the supply function should have $N$ \textit{discrete} outputs so that the bidding power can be expressed with the $N$-dimensional power bids.

\begin{figure}[h]
    \centering
    \includegraphics[width=0.9\linewidth]{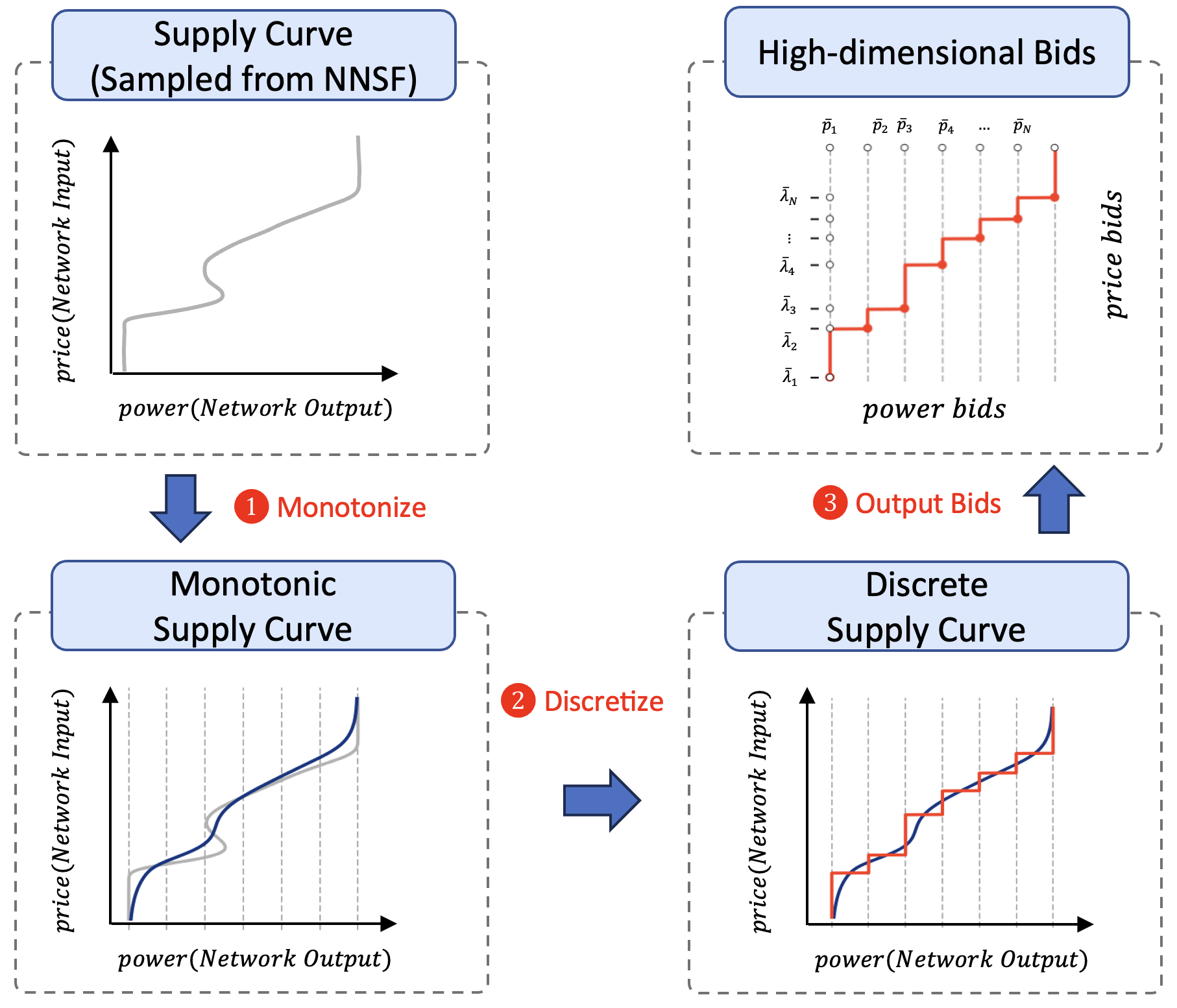}
    \caption{Represent the high-dimensional bids with monotonic and discrete-output neural networks}
    \label{fig:bid_fact}
\end{figure}

As shown in \cref{fig:bid_fact}. We propose a three-step process to generate HDBs from supply curves. First, we need to monotonize the supply curve so that the supply curve satisfies the monotonicity requirement. Second, we need to discretize the network output values so that the supply curve satisfies the discrete-output requirement. Finally, because the monotonized and discretized neural network uniquely corresponds to an HDB, the HDB parameters can be extracted.
Next, we will implement an algorithm to realize the HDB extraction process.

In the following contexts, we define the parameters of HDBs by price-power anchor points: $[\bar{\lambda}_i,\bar{p}_i]$, where $i=1,...,N$. According to the HDB rule \cref{energy_bid_rule} , the bid price and bid power should be monotonically increasing: $\bar{\lambda}_i\le \bar{\lambda}_{i+1},\bar{p}_{i}\le \bar{p}_{i+1}$.

We will shorthand the NNSF as $p=\pi(\lambda)$ to show its nature as a supply function. We will denote the supply curve as $(\lambda_k,p_k)$ , where $k=1,...,M$, to show its nature as a point set. Note that $p_k = \pi(\lambda_k)$

The HDB extraction problem is to approximate the supply curve $(\lambda_k,p_k)$ with $[\bar{\lambda}_i,\bar{p}_i]$ with low approximation errors.

The first step is to monotonize the supply curve. In the experiment sections, we will show that the supply curves can be considered naturally monotonic in most cases. Therefore, the monotonize step is not necessary most of the time.
For the non-monotonic cases, we use a simple method to monotonize the supply curve: The sampled supply curve points are valued as the cumulative max:  $p_k\gets\max\{p_1,p_2,...,p_k\},~\forall k=[1,M]$. So that $p_{k-1}\le p_k$, and the supply curve is monotonic.

The second step is to discretize the supply curve. In this step, we find the optimal HDB parameters to approximate the monotonized supply curve using an iterative algorithm.
We define the approximation error as the squared deviation between the original NNSF and the approximated HDB. The approximation error is denoted as $e_d$. It can be written as the sum of the approximation errors of each HDB segment to the supply curve:

\begin{equation}
    e_{d} = \sum_i \int_{\bar{\lambda}_i}^{\bar{\lambda}_{i+1}}\left(\pi(\lambda)-\bar{p}_i\right)^2 d \lambda
\end{equation}

Then, to minimize the error, we propose an iterative \textit{greedy algorithm}. In each iteration, we find the optimal value of $\bar{\lambda}_{i}$ and $\bar{p}_{i}$ given other $\bar{\lambda}_{-i}$ and $\bar{p}_{-i}$ values fixed,  where $-i$ means all other indexes except for $i$.
% In other words, in each iteration, we find the optimal  $\bar{\lambda}_{i}$ and $\bar{p}_{i}$ under the assumption that other   $\bar{\lambda}_{-i}$ and $\bar{p}_{-i}$ are fixed,
Using optimality conditions, at the optimal $\bar{\lambda}_{i}$ and $\bar{p}_{i}$ , the partial derivatives of $\bar{\lambda}_{i}$ and $\bar{p}_{i}$ with respect to $e_d$ should be zero:

\begin{equation}
    \begin{aligned}
0=\frac{\partial e_d}{\partial \bar{\lambda}_i} & =\frac{\partial}{\partial \bar{\lambda}_{i}}\left[\int_{\bar{\lambda}_{i-1}}^{\bar{\lambda}_{i}}\left(\pi(\lambda)-\bar{p}_{i-1}\right)^2 d\lambda \right.\\
&~ \left. +\int_{\bar{\lambda}_{i}}^{\bar{\lambda}_{i+1}}\left(\pi(\lambda)-\bar{p}_{i}\right)^2 d \lambda\right] \\
& = \left(\pi(\bar{\lambda}_{i})-\bar{p}_{i-1}\right)^2-\left(\pi\left(\bar{\lambda}_{i}\right)-\bar{p}_{i}\right)^2
\end{aligned}
\end{equation}
\begin{equation}
    \begin{aligned}
0=\frac{\partial e_d}{\partial \bar{p}_i} & =\int_{\bar{\lambda}_i}^{\bar{\lambda}_{i+1}} \frac{\partial}{\partial \bar{p}_i}\left(\pi(\lambda)-\bar{p}_i\right)^2 d \lambda \\
& =2\left[\int_{\bar{\lambda}_i}^{\bar{\lambda}_{i+1}} \pi(\lambda) d \lambda-\bar{p}_i\left(\bar{\lambda}_{i+1}-\bar{\lambda}_i\right)\right]
\end{aligned}
\end{equation}
The optimality conditions can be simplified using the monotonic feature of HDBs to:

\begin{equation}
    \left\{\begin{array}{l}
\bar{\lambda}_i=\pi^{-1}\left(\frac{1}{2}\left(\bar{p}_{i-1}+\bar{p}_i\right)\right) \\
\bar{p}_i=\int_{\bar{\lambda}_i}^{\bar{\lambda}_{i+1}} \pi(\lambda) d \lambda /\left(\bar{\lambda}_{i+1}-\bar{\lambda}_i\right)
\end{array}\right.
\label{equ:bid_extract_1}
\end{equation}
where $\pi^{-1}$ is the inverse function of $\pi$. \cref{equ:bid_extract_1} calculates the optimal  $\bar{\lambda}_{i}$ and $\bar{p}_{i}$ explicitly. However,  because the inverse function and the integration of $\pi(\lambda)$ cannot be precisely calculated, we use supply curves to approximate these two terms so that \eqref{equ:bid_extract_1} can be computed.

To compute \eqref{equ:bid_extract_1}, first, we use the supply curve to approximate the inverse function and the integrations. The inverse function can be achieved by looking up the value of the supply curve where $p_k=\frac{1}{2}\left(\bar{p}_{i-1}+\bar{p}_i\right)$, and returing the corresponding $\lambda_k$. If the supply curve samples $p_k$ cannot exactly match $\frac{1}{2}\left(\bar{p}_{i-1}+\bar{p}_i\right)$, the closest $p_k$ is chosen. Next, to approximate $\bar{p}_i$, the formula  \eqref{equ:bid_extract_1} can be understood as the mean value of $\pi(\lambda)$ in the range $\bar{\lambda}_i$ to  $\bar{\lambda}_{i+1}$, which can be approximated with the mean value of $\{p_k\}$ in the range $\bar{\lambda}_i$ to  $\bar{\lambda}_{i+1}$. Finally, \eqref{equ:bid_extract_1} can be approximately computed as:

\begin{equation}\label{equ:bid_extarct_2}
          \left\{\begin{array}{l}
        \bar{\lambda}_i=\lambda_k, \text { where } p_k=\frac{1}{2}\left(\bar{p}_{i-1}+\bar{p}_i\right) \\
        \bar{p}_i=\operatorname{mean}\left( \{p_k\}\right), \text { for all }k: \bar{\lambda}_i \leqslant \lambda_k < \bar{\lambda}_{i+1}
        \end{array}\right.  
\end{equation}

\eqref{equ:bid_extarct_2} represents the fundamental iterative step of the HDB discretization algorithm. All anchor points $[\bar{\lambda}_i,\bar{p}_i]$ are updated accordingly until convergence. 

\color{black}%was black
The output of the final step of \cref{equ:bid_extarct_2} will satisfy the requirements of a market HDB. 
First, because the supply curve is monotonized in the first step, the generated HDBs will satisfy the monotonization requirement.
Also, because the output value $\bar{\lambda}_i,\bar{p}_i$ corresponds to the discrete steps of HDBs, the discretization requirement is also satisfied.
As a result,  $[\bar{\lambda}_i,\bar{p}_i]$ can be extracted and output as the HDB and output as a valid HDB.
\color{black}
% In practice, the first and last anchor points in \cref{equ:bid_extract_1} require special treatment. For iteration on the first anchor point (i.e., $(x^1_p,y^1_p)$), an auxiliary point $(\lambda_{\min},p_{\min})$ is added to its left to indicate cases of minimum power output. Similarly, for iteration on the last anchor point (i.e., $(x^{-1}_p,y^{-1}_p)$), an auxiliary point $(\lambda_{\max},p_{\max})$ is added to its right to indicate cases of maximum power output.

% Additionally, from the applicational point of view, special operating points of some market subjects need to be met. Such as the idle state of energy storage($p=0$). 
% To meet these special operating setpoints, we add an additional phase to the algorithm. In the initial stage, an HDB is formed through \cref{equ:bid_extract_1}, disregarding special operating points. Then, the power values of specific operating points $p_k^*$, and the closest pinpoints $\bar{p}_i$ to $p_k^*$ are identified. We fix $\bar{p}_i \equiv p_k^*$ and proceed the second phase iteration. Thus, the special operational points are satisfied without significantly affecting the optimality of HDB under $e_d$.

\color{black}
\subsection{The proposed HDB generation Algorithm} \label{sec:HDB-generation-alg}

The pseudo-code of the HDB generation algorithm combining supply curve sampling and HDB extraction is shown in \cref{alg:bid_discretization}. The HDB generation precision and computational efficiencies will be discussed in the experiment sections.

\begin{algorithm}[h]
	\caption{ HDB generation}
         \label{alg:bid_discretization}
	\begin{algorithmic}[1]
        \color{black}
        \State \textbf{Input}: The NNSF $\pi$ with trained parameter $\theta $.
        \State \textbf{Otput}: High-dimensional Bids: $\{(\lambda_t^i, p_t^i)\}$
\For{t = 1, 2, ..., T} \Comment{For $T$ periods that requires HDB generation.}
    \State Get the market observation $\{\hat{s}_t\}$.
    \Statex \# Supply Curve Sampling
    \State Compose the full observaion with price ranges $(\hat{s}_t,\lambda_k)$ for $\lambda_k=\lambda_{\min},...,\lambda_{\max}$ with stepsize $\delta \lambda$.
    \State Parallelly Sample the NNSF by  $p_k=\pi_\theta(s_t,\lambda_{k})$ for supply function point set $(\lambda_k,p_k)$
    \Statex \# HDB Extraction
    \State $p_i\gets\max\{p_1,p_2,...,p_i\}$  \Comment{ a. Monotonize}
    \State Initialize $(\bar{\lambda}_{i},\bar{p}_{i}), i\in\{1,2,..,P\}$ \Comment{b. Discretize}
    \State $(\bar{\lambda}_0,\bar{p}_0) \gets (\lambda_{\min},p_{\min})$
    \State $(\bar{\lambda}_{P+1},\bar{p}_{P+1}) \gets (\lambda_{\max},p_{\max})$
    \While{not converge}
    \For{$i\in \{1,2,..,P\}$}
        \State Update each pair $(\bar{\lambda}_i,\bar{p}_i)$ according to:
        \begin{align*}
          \left\{\begin{array}{l}
        \bar{\lambda}_i=\lambda_k, \text { where } p_k=\frac{1}{2}\left(\bar{p}_{i-1}+\bar{p}_i\right) \\
        \bar{p}_i=\operatorname{mean}\left( \{p_k\}\right), \text { for all }k: \bar{\lambda}_i \leqslant \lambda_k < \bar{\lambda}_{i+1}
        \end{array}\right.  
        \end{align*}
    \EndFor
    \EndWhile

    \State Store the HDB: $(\bar{\lambda}_i, \bar{p}_i)$, $i=1,...,N$ as $\{(\lambda_t^i, p_t^i)\}$ \Comment{c. Output Bids}
    \EndFor

	\end{algorithmic} 
\end{algorithm}
\color{black}

\subsection{Discussion: the Difficulty of Applying RL to the HDB generation framework } \label{sec:RL-suitability}

Note that, even though the proposed HDB generation framework(\cref{fig:generation_framework}) is able to generate HDBs from NNSFs, it can hardly be used to train an NNSF neural network with RL.

The main reason is that the HDB generation framework does not follow the typical state-\textit{action}-reward structure in RL. Instead, it is a state-\textit{multiple\_actions}-reward structure.
To be specific, because the final bid (the submitted HDB) is extracted from the NNSF's output across the whole price range, the NNSF is forwarded $M$ times (usually $M$ is hundreds of times to generate a supply curve). Because one reward corresponds to hundreds of actions, it is impractical to optimize the NNSF with such an action-reward ratio.

To tackle this problem, in the next section, we will approximate the HDB generation framework to be more RL-friendly and propose an RL-based NNSF training algorithm. Further, we will demonstrate the effectiveness of the proposed approximation in the experiments.

\section{RL-based NNSF Training}\label{sec:HDBlearn}

In this section, we will propose an approximation of the HDB generation framework (\cref{fig:generation_framework}) and use the approximation to train NNSFs with RL algorithms (\cref{sec:ppo_rl_algorithm}). Finally, a refinement of the NNSF output space is proposed for ESS bidding in \cref{sec:action-space-refinement}.

\subsection{Simplified Bidding Framework for RL}\label{sec:learning-framework}

In this subsection, we first simplify the HDB generation framework by approximating the market clearing result with NNSF's output. Then, we propose a more RL-friendly bidding process that is equivalent to HDB bidding. It will follow the typical state-action-reward RL structure.

\begin{figure}[h]
    \centering
    \includegraphics[width=1\linewidth]{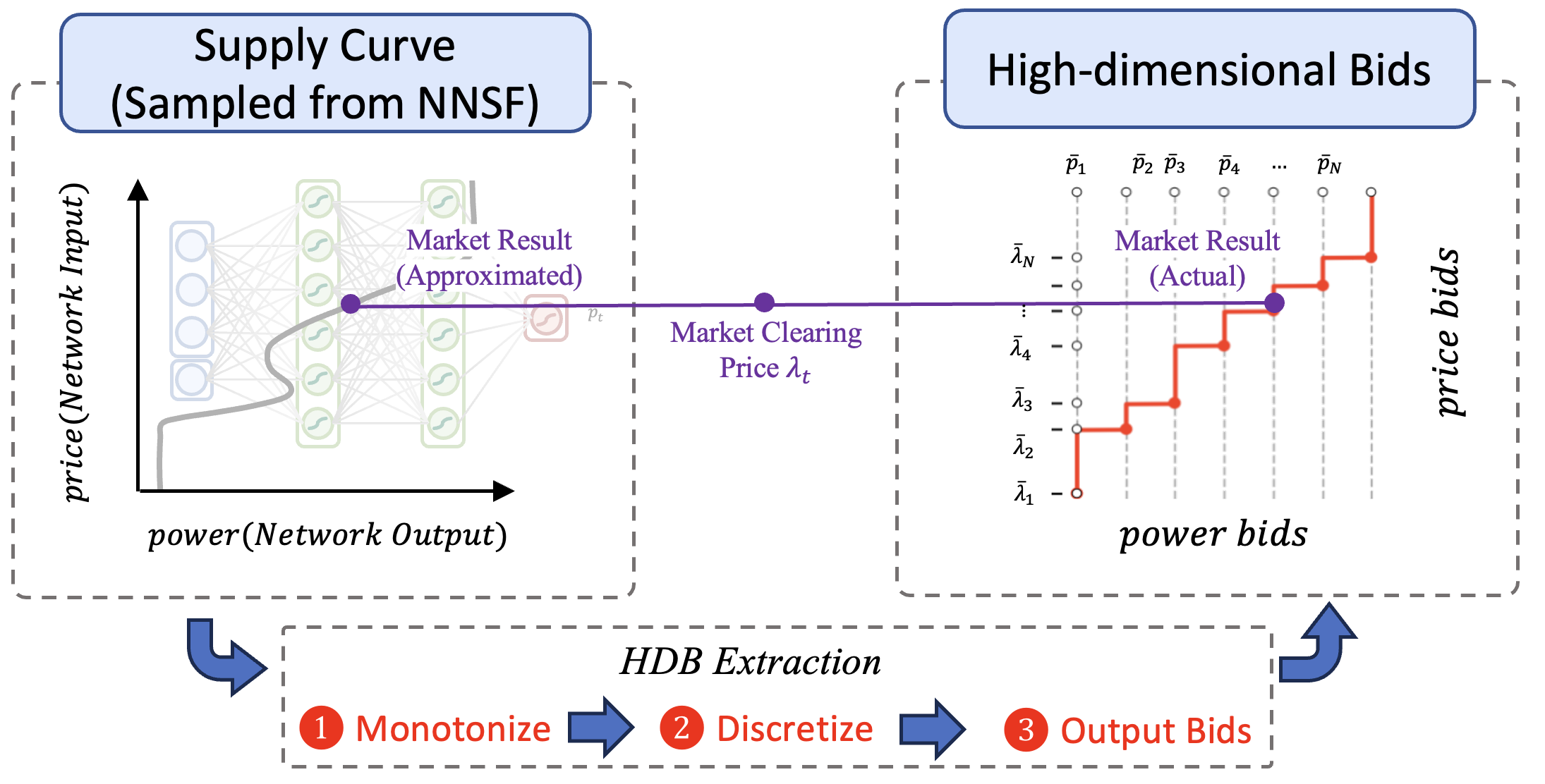}
    \caption{Approximate the market clearing result with NNSF's output at the market clearing price}
    \label{fig:market_clearing_approximation}
\end{figure}

\cref{fig:market_clearing_approximation} shows how market clearing results can be approximated with the NNSF's output at the market clearing price. Recall that the actual market clearing result is a \textit{point sample} on the HDB bidding curve (right purple dot). It is the HDB supply function's value at the market clearing price. Because the HDB supply function is an approximation of the supply curve. The market clearing power can be approximated with the supply curve's value at the market clearing price.

\color{black}
In other words, to get the market clearing result, we do not need to formulate the full HDB by supply curve sampling and HDB extraction. Instead, we can approximate the market clearing result by the output of the NNSF at the \textit{actual market clearing price}.
\color{black}

\begin{figure}[h]
    \centering
    \includegraphics[width=1\linewidth]{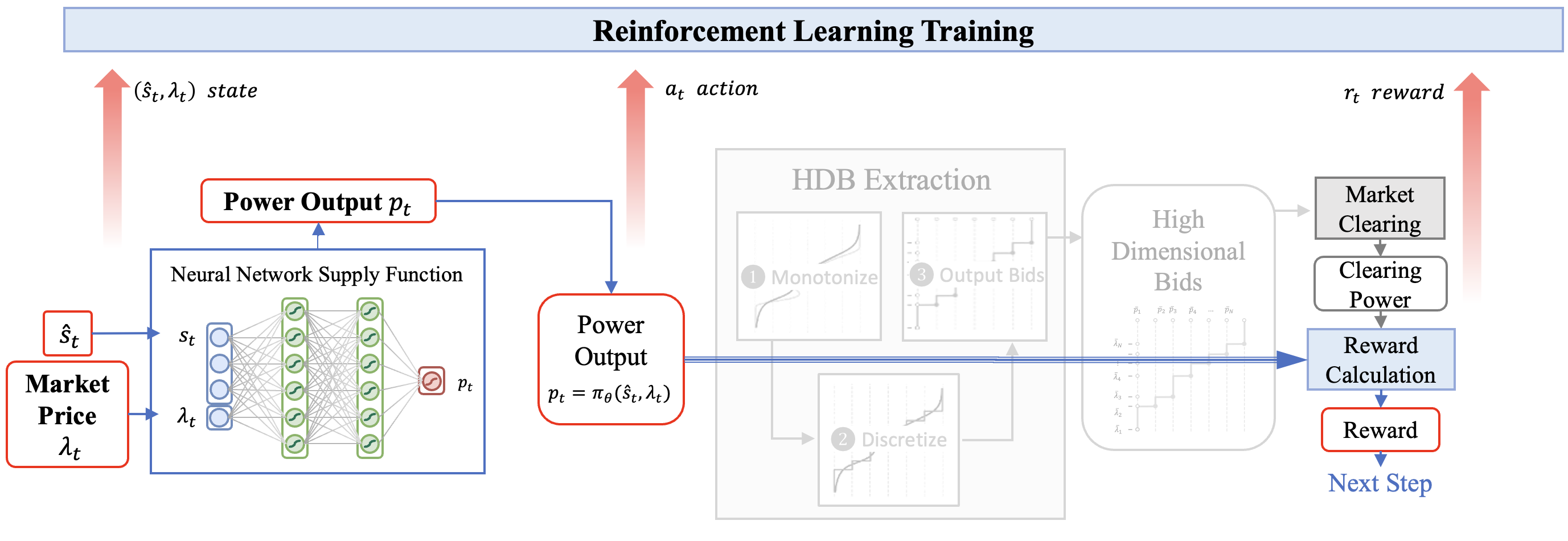}
    \caption{The proposed NNSF training framework}
    \label{fig:learning_framework}
\end{figure}

Leveraging such an approximation, the HDB generation framework (\cref{fig:generation_framework}) can be simplified to the NNSF training framework (\cref{fig:learning_framework}).
In each step, the market state $\hat{s}_t$ and market clearing price $\lambda_t$ are observed. They are combined to be state $s_t=(\hat{s}_t,\lambda_t)$.  $s_t$ is used to sample the NNSF's output of the market clearing price. The output power value denoted as $p_t$. $p_t$ is considered to be the approximate result of the extracted HDB, and the bidding profit of  $p_t$ is computed as the bidding reward.
Finally, the bidding history $(s_t, a_t, r_t, s_{t+1})$ is collected for RL training, and the bidding process proceeds to the next step.

Such a NNSF training framework (\cref{fig:learning_framework}) is compatible with RL algorithms.
It is a typical state-action-reward MDP. The state is the market information $\hat{s}_t$ and the market clearing price $\lambda_t$, the action is the power output $p_t$, and the reward is the market bidding profit \cref{equ:ESS_reward}.

Note that the training framework has a low-dimensional action space, which is the power output $p_t$. It transforms the high-dimensional bidding problem into a low-dimensional power dispatch problem. From the perspective of training an RL policy, the proposed RL problem follows the typical state-action-reward structure and can be learned more easily.

Nonetheless, the NNSF training framework does not have the exact same outcome as the HDB generation framework. The NNSF training framework uses the NNSF's output to approximate the HDB generation and market clearing process. It will introduce approximation errors. In the experiments, we will show that such errors can be neglected in practice.

Next, we will propose an RL-based NNSF training algorithm based on the proposed training framework.

\subsection{NNSF Training with the PPO RL Algorithm}\label{sec:ppo_rl_algorithm}

\color{black}
This subsection trains the NNSF with the well-known PPO RL algorithm\cite{schulman_proximal_2017}. Other RL algorithms with continuous action spaces are also applicable to our framework. Recall that an RL algorithm improves a policy network by interacting with the environment, and it updates the policy with the interaction history.

We will first introduce the basic principles of RL and the PPO RL algorithm.
The PPO algorithm is an actor-critic style RL algorithm. Actor-critic style RL algorithms learn a policy function $\pi_\theta(a_t|s_t)$ and a value function $V_\phi(s_t)$ at the same time.

The policy function $\pi_\theta$ is a mapping from state to action distribution. It maps the state $s_t$ to an action distribution $\pi_\theta(a_t|s_t)$, where $\theta$ are the policy parameters. The action distribution is defined by the neural network’s output, which consists of the mean and variance parameters of a Gaussian distribution.
The value function $V_\phi(s_t)$ is a mapping from state to state's value. It estimates the expected return of a state $s_t$, which is $V_\phi(s_t)=\mathbb{E}\{\Sigma_{t=t}^\infty \gamma^tr_t(s_t,a_t)\}$. $\phi$ represents the network parameters of the value function. 

The value function $V_\phi(s_t)$ can be learned by minimizing the error to the estimated value $\hat{R}_t$: 

\begin{equation}
\label{equ:value-function-td}
    \phi_{k+1}=\arg \min _\phi  \sum_{\tau \in \mathcal{D}_k} \sum_{t=0}^T\left(V_\phi\left(s_t\right)-\hat{R}_t\right)^2
\end{equation}
where $\hat R_t = r(s_t,a_t) + \gamma V_\phi(s_{t+1})$ is the temporal difference target of  $V_\phi\left(s_t\right)$ and $\mathcal{D}_k$ is the rollout buffer that stores system transitions $\tau = (s_t,a_t,r_t,s_{t+1})$.

Likewise, the policy function $\pi_\theta(a_t|s_t)$ can be learned by policy gradient methods. $\pi_\theta(a_t|s_t)$ represents the probability distribution of actions $a_t$ under state $s_t$. i.e. $a_t\sim \pi_\theta (s_t)$. Policy gradient methods estimate the gradient of the policy parameters $\theta$ to maximize the following objective:
\begin{equation}\label{equ:original-PG-objective}
    \theta_{k+1}=\arg \max _\theta \sum_{\tau \in \mathcal{D}_k} \sum_{t=0}^T \frac{\pi_\theta\left(a_t \mid s_t\right)}{\pi_{\theta_k}\left(a_t \mid s_t\right)} A^{\pi_{\theta_k}}\left(s_t, a_t\right)
\end{equation}
where $\pi_\theta\left(a_t \mid s_t\right) $ and $ \pi_{\theta_k}\left(a_t \mid s_t\right)$ are the probabilities of taking action $a_t$ under $\theta$ and $\theta_k$(policy parameters used to sample the $k$th rollout buffer). $A^{\pi_{\theta_k}}\left(s_t, a_t\right)$ is an estimate of the \textit{advantage} that can be gained under policy $\pi_{\theta_k}$ from taking action $a_t$ rather than following action distribution $a\sim\pi_{\theta_k}$. $A^{\pi_{\theta_k}}\left(s_t, a_t\right)$ is derived based on the trajectory $\tau$ and value function $V_\phi$\cite{schulman_high-dimensional_2018}. This iterative step maximizes the gained advantage of the updated policy parameter $\theta_k\rightarrow\theta_{k+1}$. 

To avoid the policy function from changing too far and causing performance collapse, PPO algorithm\cite{schulman_proximal_2017} adopts a new surrogated policy gradient objective rather than \cref{equ:original-PG-objective}:

\begin{align}
\label{equ:PPO-policy-gradient}
    \theta_{k+1} = \arg \max _\theta  \sum_{\tau \in \mathcal{D}_k} \sum_{t=0}^T &\min \left(\frac{\pi_\theta\left(a_t \mid s_t\right)}{\pi_{\theta_k}\left(a_t \mid s_t\right)} A^{\pi_{\theta_k}}\left(s_t, a_t\right), \right. \nonumber \\
    &\quad \left. \vphantom{\frac{\pi_\theta\left(a_t \mid s_t\right)}{\pi_{\theta_k}\left(a_t \mid s_t\right)}} g\left(\epsilon, A^{\pi_{\theta_k}}\left(s_t, a_t\right)\right)\right)
\end{align}

where $g(\cdot)$ scales $A$ according to $\epsilon$:

\begin{equation}
    g(\epsilon, A)= \begin{cases}(1+\epsilon) A & A \geq 0 \\ (1-\epsilon) A & A<0\end{cases}
\end{equation}

This objective makes sure that the updated policy stays close to the original policy.  So that the policy won't change too fast to cause performance collapse. For the detailed explanations, please refer to \cite{schulman_proximal_2017}.
\color{black}

We can combine the proposed NNSF training framework(\cref{fig:learning_framework}) with the PPO RL algorithm.
The PPO-based NNSF training approach is summarized in  \cref{alg:PPO}.

\begin{algorithm}
	\caption{PPO-based NNSF Training}
        \label{alg:PPO}
	\begin{algorithmic}[1]
    \State Input: initial NNSF parameters $\theta_0$, initial value function parameters $\phi_0$
		\For {$k=0,1,2, \ldots, N$} \Comment{For $N$ rollouts}

            \For {$t=0,1,2, \ldots, T$} \Comment{For $T$ timesteps}
            \State Get market state $\hat{s}_t$, and the market clearing price $\lambda_t$.
            \State Merge market state and market price as the state: $s_t=(\hat{s}_t,\lambda_t)$.
            \State Sample the approximate market clearing result $p_t$ from the NNSF's output distribution $p_t\sim\pi_\theta(s_t)$.
			\State Compute the reward $r_t$ by market profit \cref{equ:ESS_reward}.
            \State Transit to the next step.
            \State Store the step history to the rollout buffer $\mathcal{D}\gets (s_t, p_t, r_t, s_{t+1})$.
            \EndFor
            \State Use the rollout buffer to update the function estimates.
            \State Compute advantage estimates, $\hat{A}_t$ using \cite{schulman_high-dimensional_2018} based on the current value function $V_{\phi_k}$.
            \State Update the policy $\pi_\theta$ by maximizing the PPO objective of \cite{schulman_proximal_2017}.
            % \cref{equ:PPO-policy-gradient}.
            \State Fit value function $V_\phi$ by regression on Value iteration of \cite{schulman_proximal_2017}.
            % \cref{equ:value-function-td}.
            
            \State Empty the rollout buffer $\mathcal{D}$.
		\EndFor
	\end{algorithmic} 
\end{algorithm}

% In \cref{alg:PPO}, to consider \textit{information legitimacy}, the market information input $\hat{s}_t$ should \textit{not} be later than the \textit{bid submission time}. Though the market clearing price $\lambda_t$ is the actual market clearing price(which is released later than the bid submission time), other input observations should not be later than the bid submission time. Because they will be used to generate HDBs for bid submission.

Because the neural network structure is not the focus of this paper.  In practice, we use common multi-layer perceptron networks that have two hidden layers of 256 nodes as the NNSF network $\pi_\theta$ and value network $V_\phi$. All output values are mapped through Tanh activations, which maps the output value from $(-\infty,\infty)$ to $(-1,1)$. Other training parameters will be detailed in the experiment section. 

\subsection{NNSF Refinement for Energy Storage Bidding} \label{sec:action-space-refinement}

In this subsection, we refine the NNSF's output space for ESS bidding.

The original output space of NNSF is the power action $p_t$, which is a continuous space between -1 and 1 (corresponding to maximum charging and discharging). This output space has difficulties in achieving zero output. \color{black}
Zero output means the ESS's charging and discharging power are both zero. It is a common decision of ESSs, and it helps the ESS to hold the stored energy for more valuable dispatches.

However, it is difficult for neural networks to stay at exactly zero output for a range of input prices. Because neural networks' inside computations are mostly smooth, it can be difficult for neural networks to achieve a full-zero output range.
\color{black}

To assist neural networks to output zero easily, we modify the output space of an NNSF to a 4-dimensional space. It includes the price boundaries for the zero output: $(\lambda^d_t, \lambda^c_t)$, and the power actions beyond the zero output price boundaries: $(p^d_t, p^c_t)$. The final power action of an NNSF is:

\begin{equation}\label{equ:nnsf4}
p_t= \begin{cases}p_t^d & , \text { if } \lambda_t \geqslant \lambda_t^d \text { (discharging) } \\ -p_t^c & , \text { else if } \lambda_t \leqslant \lambda_t^c \text { (charging) } \\ 0 & , \text { else }\end{cases}
\end{equation}

which assures the power action $p_t$ is zero in the range $(\lambda^c_t,\lambda^d_t)$, so that it is easier for neural networks to achieve zero output for a specific price range. The effectiveness of such refinement will be demonstrated in the experiments.

\section{Performance Evaluation}
\label{sec:experiments}

In this section, we demonstrate the proposed method's effectiveness in training NNSF and generating HDBs on various market price nodes and different ESS parameters.
We also compare it against other existing RL-based bidding methods that use LDBs. 
% The advancement of the proposed method is demonstrated by the comparisons.

\subsection{Experiment Setup}

We consider a real-world ESS in the RT energy market\cite{PJM-Energy-and-Ancillary-Service-Market-Operations}. As an ESS usually has a small portion of the market, we assume it to be a price-taker. It maximizes its profit with HDB bidding\cref{energy_bid_rule}.

The ESS has $\pm$1MW power capacity and different storage capacities ranging from 2MWh to 12MWh. Its charging and discharging efficiency are both $0.95$. Its degradation cost  $\lambda^{dep}$ is 10USD/MWh, which means a 1MWH energy cycle has a cost of about 10USD.

\relax{The PJM market is a real-world power market that distributes electricity across several states in the Eastern and Midwestern US. \color{black}
Based on the submitted market bids, PJM organizes the buying and selling of electrical energy for 12$\times$five-minute intervals of the operating hour. 
\color{black} Then, 12 clearing prices and 12 power setpoints are fed back to the bidders as the market clearing result.\color{black}

% Data source
\color{black}
The RT market data are collected from the PJM market history. 
We select five price nodes of the PJM market as the data source, including PJM-RTO (Cental Node), DOM (SouthEast), EKPC (SouthWest), COMED (NorthWest), and PSEG (NorthWest). \color{black}
The PJM RT market prices from 2018/4/1-2020/12/25 (1000days) are used to simulate the RT market. \color{black}
The PJM RT market data is split into train and test sets.\color{black}
 The first 70\% market data are used for training using \cref{alg:PPO} and the last 30\% market data are used for testing using \cref{alg:bid_discretization}.

% observation setup
In order to effectively arbitrage under high uncertainty, the following observations are provided for the ESS bidder: 1) bid time, denoted as $T_t$, which is $sin(T/24)$,$cos(T/24)$. It is a 2-dimensional sinusoidal encoded time to indicate the time of day. 2) market histories, denoted as $h_t$.
It consists of the angle and amplitude of the first three dimensions in the discrete Fourier Transform of the past 6 hours' RT market price and the past 4 days' DA market price. 3) energy levels, denoted as $e_t$, which is the SoC of the ESS.

% training parameters
All experiments were run on a PC with an i9-10900X CPU and 128GB memory. The RL algorithms are trained using an NVIDIA 3090 GPU with 24GB memory. 
We implement the proposed RL algorithm in PyTorch. The detailed hyperparameters are shown in \cref{tab:hyperapram}:

% \usepackage{graphicx}
\begin{table}[h]

\caption{RL Algorithm Hyperparameters}
\label{tab:hyperapram}

\resizebox{\columnwidth}{!}{%
\begin{tabular}{cc|cc}
\hline
Parameter          & Value                 & Parameter            & Value            \\ \hline
$N$& 10& $M$& 512\\
Total Steps        & $3\times 10^6$        & Batch Size           & 256              \\
Action Initial Std & 0.6                   & $\epsilon_{clip}$    & 0.2              \\
Action Final Std   & 0.25                  & $\gamma$             & 0.999            \\
Action Std Decay   & $2\times10^{-7}$/step & SoC Violate Penalty $P$ & 170 USD \\
 Actor Learning Rate  & $5\times10^{-5}$ & $\lambda_{\min}$&-50 USD\\ 
 Critic Learning Rate & $3\times10^{-4}$ & $\lambda_{\max}$&200 USD\\\hline
\end{tabular}%
}

\end{table}

\subsection{Benchmark Methods}

To verify the effectiveness of the proposed HDB bidding method, it is tested against other RL-based bidders with LDBs.

\begin{enumerate}
    \item Self-Scheduling Bid Learning (Self-Bid)\cite{wei_self-dispatch_2022}: 
    The Self-Bid method\cite{wei_self-dispatch_2022} only bids one power quantity $p_t$, and the power quantity is always accepted by the market.
    % \color{black} It is a price taker of the market, and the market clearing price is paid to the bidder. This is the simplest yet the most used bid format in the current literature. \color{black}
    \item Two Pair Bid Learning (Pair-Bid)\cite{tao_deep_2022}: 
    The Pair-Bid method\cite{tao_deep_2022} bids two price-power pairs. One bid is used for discharging power, and another pair is used for charging power (energy storage only). The power bid $p_t$ is accepted when the market clearing price is higher/lower than the price bid depending on the charging or discharging direction. 
    % \color{black}Pair-Bid is more flexible than Self-Bid because it can respond to market prices with price thresholds.\color{black} 
    We consider Pair-Bid to represent the state-of-the-art performance of LDBs.
    \item Direct HDB Bid Learning (Direct-HDB):
    Direct-HDB is an upgraded version of Pair-Bid, where $N$ instead of 2 pairs are output by the policy for market bidding. This is the most direct approach to generate HDBs with neural networks, which includes an RL policy with a 2$N$-dimensional output. The first $N$ dimensions are used as price bids, and the next $N$ dimensions are used as power bids. 
    It serves as a baseline to verify the effectiveness of the proposed HDB bidding method.
    \item HDB Bidding (HDB-Bid): The bidding method proposed in this paper. 
    \color{black}The NNSF is trained using RL (\cref{alg:PPO}), and the HDB generation process (\cref{alg:bid_discretization}) is used to generate test results. The NNSF's action space is refined as in \cref{sec:action-space-refinement}\color{black}
    \item HDB Bidding without action space refinement (HDB-WOA): The NNSF training and HDB generation process of HDB-WOA is the same as the HDB-Bid method. However, its action space is not refined and is a one-dimensional power output.
    \item Optimal Bidding (Optimal-Bid): Optimal-Bid is the best possible bidder. 
    % It cannot be achieved in reality but can be solved afterward based on price histories. 
    It is the upper-bound performance of any actual bidder. 
    % Optimal-Bid is used to measure the performance of other methods. 
    In the following context, the bidding results will be scaled based on Optimal-Bid to a percentage representation.
\end{enumerate}

\subsection{General Performance}
% 训练曲线

In this subsection, we compare the general performance of the proposed HDB-Bid bidding method with Self-Bid, Pair-Bid, Direct-HDB, and HDB-WOA. \color{black}
The performance metrics include the training curve, the culmulative test reward and the captured profit ratio. The RL-based ESS bidders are trained for $2.5\times 10^7$ steps in the simulated environment. Each takes about $2$ hours on the mentioned hardware. They are tested on unseen datasets of length 300 days. \color{black}

\subsubsection{Training Curves on the Train Dataset}
The training curves of different RL-based bidding methods is shown in \cref{fig:train_curve}. To ensure a fair comparison between different bid formats, the PPO training algorithm with the same hyperparameter is used for training. Each figure shows the training curves at of specific energy storage capacity. The shown training curves are the average training reward across five price nodes, and the shallow areas in the background are the maximum and minimum training rewards across the five price nodes.

\begin{figure}[h]
    \centering
    \includegraphics[width=1\linewidth]{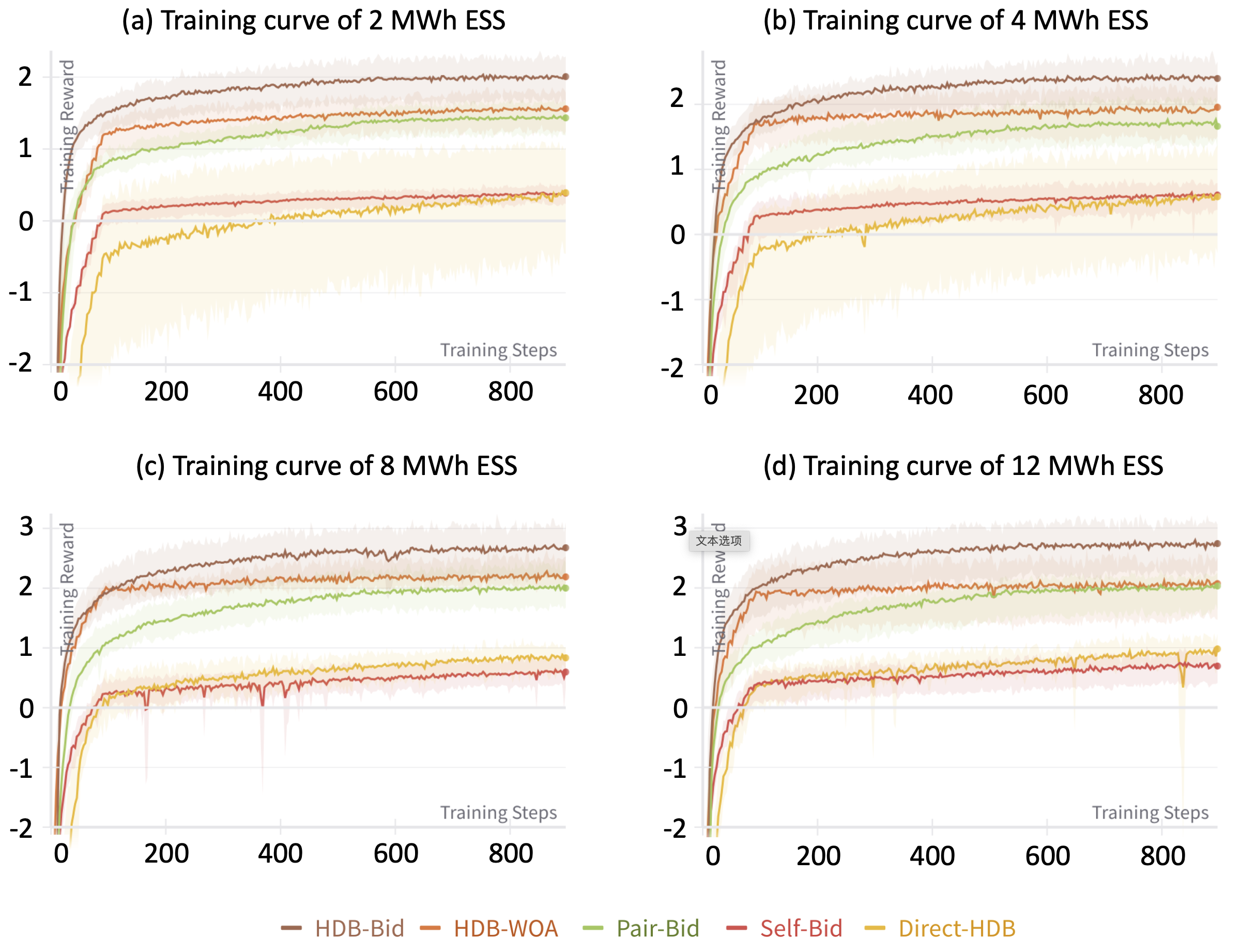}
    \caption{The training curve of different bidding methods on different energy storage capabilities. The reward values are averaged across five price nodes.}
    \label{fig:train_curve}
\end{figure}

% HDB 的整体效果最好，其他的
The proposed HDB bidding method outperforms other bidding methods in terms of training speed and performance. It achieves the fastest convergence speed and the highest convergence performance. 
\color{black}HDB-WOA has a similar convergence speed in the initial stages. However, the final performance of HDB-WOA is lower because HDB-WOA cannot maintain the NNSF output power at exactly zero, so it is not able to learn power withholding strategies and the bidding performance is affected. \color{black}
The Pair-Bid achieves a training performance similar to that of HDB-WOA but learns in a slower way. 
The Direct-HDB and Self-Bid exhibit similar training patterns. 
\color{black} They have similar training performance in low-capacity cases (2MWh and 4MWh), and Direct HDB has a higher training performance in high-capacity cases (8MWh and 12MWh).\color{black}
Their training performance is mediocre compared with the first three bidding methods.

\color{black}
% The gap between HDB-Bid and Pair-Bid demonstrates the improvements achieved by observing the market price $\lambda_t$ in the neural network input.
% Note that HDB-Bid and Pair-Bid have the same neural network output. Because the Pair-Bid format is equivalent to the output format described in \cref{equ:nnsf4}, which includes the charging/discharging power quantity and the respective price threshold. Therefore, they have the same neural network output space. Their main difference is the neural network input space. Pair-Bid only has the market state $\hat{s}_t$ as the network input, while HDB-Bid has an additional market price input $\lambda_t$. $\lambda_t$ is additional information that can be observed from our proposed bidding framework. In fact, training HDB-Bid and Pair-Bid majorly differs from this additional price input. From the training results, we can observe that observing $\lambda_t$ is beneficial for bidding under high uncertainties, which provides a 6.77\%-16.95\% training performance increase.

% 对比不同的HDB，可以发现所提出的方法的有效性
The different training performances of HDB-based methods demonstrate the effectiveness of the proposed HDB representation and bidding method. 
Direct-HDB is a direct approach to generating raw HDBs for bidding, but the training performance of Direct-HDB is similar to Self-Bid. Since the Self-Bid is self-dispatch with only one-value power output, this demonstrates that Direct-HDB deteriorates to the bidding performance of the simplest bid format. 

% The training performance of HDB-WOA is a little less than that of HDB-Bid. This is mainly due to HDB-WOA's inability to withhold the stored energy for more valuable dispatches. 
%  \cref{sec:action-space-refinement} empowers HDB-Bid with the ability to hold the power output at zero, and the effectiveness of this additional ability is shown in this comparison. Overall, the proposed HDB-Bid achieves the best bidding performance using HDBs for RL-based methods. 

By comparing the performance of different bidding formats (HDB, Pair-Bid, and Self-Bid), we can observe that more flexible bidding formats can achieve higher bidding performance. From the perspective of RL, more flexible bidding formats can provide more flexible action spaces (like the price thresholds) and more informative state spaces (like the market price). Both can improve the RL's training performance as long as they can be effectively utilized by the RL algorithms.
\color{black}

% 投标结果，介绍bar chart和表格
\subsubsection{Captured Profit Ratios on the Test Dataset}

\color{black}

\begin{table*}[h]

\caption{Bidding performance compared with optimal market income in percentage}
\label{tab:bid_performance}

\centering
\color{black}
\begin{tabular}{c|c|cccccc}
\hline
Node                     & Capacity & Optimal Profit(USD) & HDB-Bid          & HDB-WOA & Direct-HDB & Pair-Bid & Self-Bid \\ \hline
\multirow{4}{*}{PJM-RTO} & 2MWh     & 196695.76
& \textbf{81.94\%} & 62.38\% & 44.54\%    & 70.37\%  & 20.15\%  \\
                         & 4MWh     & 215900.52
& \textbf{82.46\%} & 63.51\% & 55.52\%    & 71.04\%  & 26.70\%  \\
                         & 8MWh     & 227115.32
& \textbf{86.99\%} & 61.26\% & 41.10\%    & 74.10\%  & 30.66\%  \\
                         & 12MWh    & 229876.92
& \textbf{85.08\%} & 63.05\% & 33.22\%    & 71.23\%  & 24.68\%  \\ \hline
\multirow{4}{*}{DOM}     & 2MWh     & 259200.00
& \textbf{83.51\%} & 58.34\% & 61.04\%    & 71.65\%  & 25.92\%  \\
                         & 4MWh     & 279523.43
& \textbf{86.16\%} & 60.79\% & 62.56\%    & 76.28\%  & 30.33\%  \\
                         & 8MWh     & 291301.26
& \textbf{87.11\%} & 60.88\% & 37.65\%    & 77.10\%  & 29.66\%  \\
                         & 12MWh    & 294945.57
& \textbf{87.44\%} & 62.17\% & 44.72\%    & 73.67\%  & 31.72\%  \\ \hline
\multirow{4}{*}{EKPC}    & 2MWh     & 210394.50
& \textbf{81.35\%} & 59.78\% & 33.26\%    & 69.00\%  & 26.87\%  \\
                         & 4MWh     & 227824.98
& \textbf{86.07\%} & 61.85\% & 39.32\%    & 69.12\%  & 29.10\%  \\
                         & 8MWh     & 237024.77
& \textbf{88.41\%} & 63.73\% & 46.32\%    & 78.82\%  & 28.95\%  \\
                         & 12MWh    & 239230.74
& \textbf{86.35\%} & 63.18\% & 42.54\%    & 78.68\%  & 29.09\%  \\ \hline
\multirow{4}{*}{PSEG}    & 2MWh     & 132436.86
& \textbf{77.49\%} & 49.08\% & 47.49\%    & 66.55\%  & 11.87\%  \\
                         & 4MWh     & 142708.82
& \textbf{79.31\%} & 46.70\% & 24.19\%    & 64.89\%  & 12.71\%  \\
                         & 8MWh     & 148976.69
& \textbf{75.26\%} & 59.22\% & 50.10\%    & 67.56\%& 11.85\%  \\
                         & 12MWh    & 151270.96
& \textbf{70.84\%} & 56.39\% & 44.28\%    & 64.07\%  & 10.79\%  \\ \hline
\multirow{4}{*}{COMED}   & 2MWh     & 218417.58
& \textbf{76.73\%} & 57.76\% & 42.79\%    & 64.91\%  & 23.74\%  \\
                         & 4MWh     & 239701.78
& \textbf{81.21\%} & 61.18\% & 25.04\%    & 71.90\%  & 29.04\%  \\
                         & 8MWh     & 251876.35
& \textbf{82.71\%} & 60.29\% & 32.39\%    & 71.47\%  & 22.89\%  \\
                         & 12MWh    & 255628.37& \textbf{82.12\%} & 60.35\% & 39.74\%    & 73.37\%  & 23.98\%  \\ \hline
 & Variance & 47717.22& 4.64\%& 4.46\%& 10.21\%& 4.32\%&6.92\%\\ \hline
\end{tabular}

\end{table*}

\color{black}

Next, testing rewards of the comparing methods for different energy storage capacities are 
listed in \cref{tab:bid_performance} and 
visualized in \cref{fig:test_rew}. 
% \cref{tab:bid_performance} groups the bidding performances by price-node and \cref{fig:test_rew} groups the plots by the energy storage capabilities.
All bidding performances are measured against the optimal market bidding result (Optimal Profit).

\begin{figure}[h]
    \centering
    \includegraphics[width=1\linewidth]{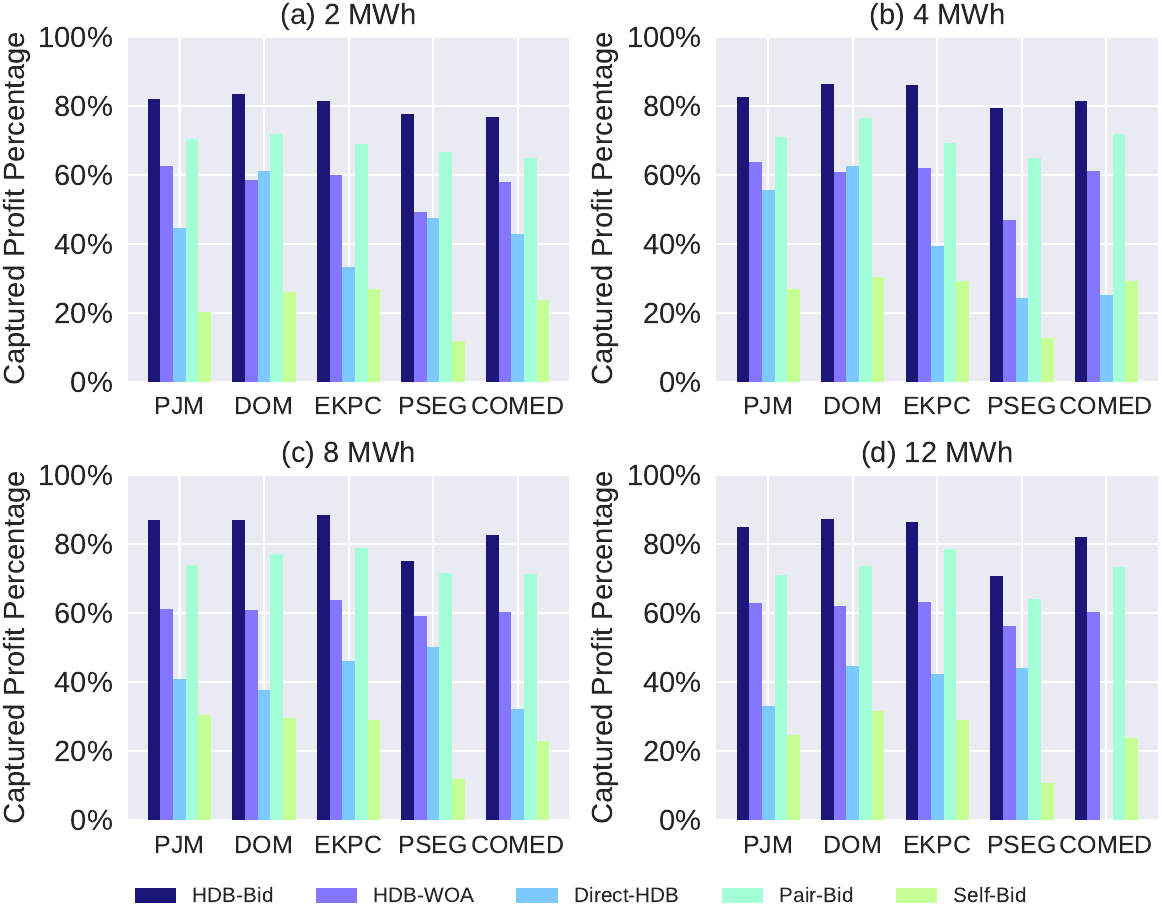}
    \caption{Captured profit percentage of different bidding methods compared with the optimal market bidding income (which is the 100\% levels of the plots)}
    \label{fig:test_rew}
\end{figure}

% 对比HDB相较其他的有什么提升
\cref{fig:test_rew} shows that HDB-Bid captures a higher percentage of the optimal market profit compared with other low-dimensional bid formats. 
HDB bidding averagely achieves 15.40\% higher profit based on the Pair-Bid bidding method, which is a 10.94\% profit boost based on the optimal profit.  This shows that higher flexibility in bids enables better learning of bidding strategies and higher income. Self-bid has the lowest bidding reward because it cannot respond to the market price with price bids.

\color{black}
% 分析不同价区Optimal Bid收益的区别
\cref{tab:bid_performance} also demonstrates the optimal profit in different price nodes varies a lot. The five price nodes are located in the same market, but their optimal profit in the RT energy market could be doubled from PSED to DOM. Also, the captured profit percentage of RL bidder is higher for high-profit regions (DOM).
This shows that the real-time energy market is a location-sensitive market, and the choice of the installed generator can influence the bidding profit by a substantial margin. 

% % 对比SoC变化对Captured profit ratio的影响
% The energy storage capability will influence the ability of the RL agent to capture the optimal profit. In general, the low-capacity cases (2MWh and 4MWh) have a lower captured profit percentage, and high-capacity cases (8MWh and 12MWh) have a higher percentage. This is because low-capacity bidders cannot save enough energy to capture the most valuable dispatches, and may miss chances in the market. However, among the high-capacity cases, we observe that 12MWh does not have a higher captured profit percentage than 8MWh. We explain this phenomenon as the 12MWh case needs to arbitrage across multiple market days, which increases the difficulty of energy withholding strategies. As a result, it is harder to capture a high percentage of optimal profit for 12MWh ESS. Overall, the 8MWh achieves the highest captured profit ratios.

% 讨论不同投标方法的收益方差
% The standard deviation of the captured percentage of the optimal profit is shown at the bottom of \cref{tab:bid_performance}. Direct-HDB has a significantly higher variance compared with other bidding methods. If we take a closer look at the captured profit percentage of Direct-HDB, we can see that Direct-HDB's bidding profit varies significantly with the energy capacity. This shows that directly generating HDB can lead to unstable training performances.
\color{black}

% 小结投标表现
In summary, the proposed HDB-Bid method is able to capture 70.84\%-88.41\% of the optimal profit. It improves the performance by 15.40\% based on the low-dimension bids (Pair-Bid) on average. To the best of our knowledge, this is the highest reported profit ratio in the literature captured by RL-based methods for RT energy market bidding.

\color{black}
\subsubsection{Culmulative Reward on the Test Dataset}
\cref{fig:culmulative_reward} shows the cumulative test reward of different bidding methods at the PJM-RTO price node. Due to the limited space, the other four price nodes are not plotted. The horizontal axis is the timestamp, which includes 300 days of unseen data. The vertical  axis is the cumulative profit of a 1MW ESS with different energy storage capabilities ranging from 2MWh, 4MWh, 8MWh, to 12MWh.

\begin{figure}[h]
    \centering
    \includegraphics[width=1\linewidth]{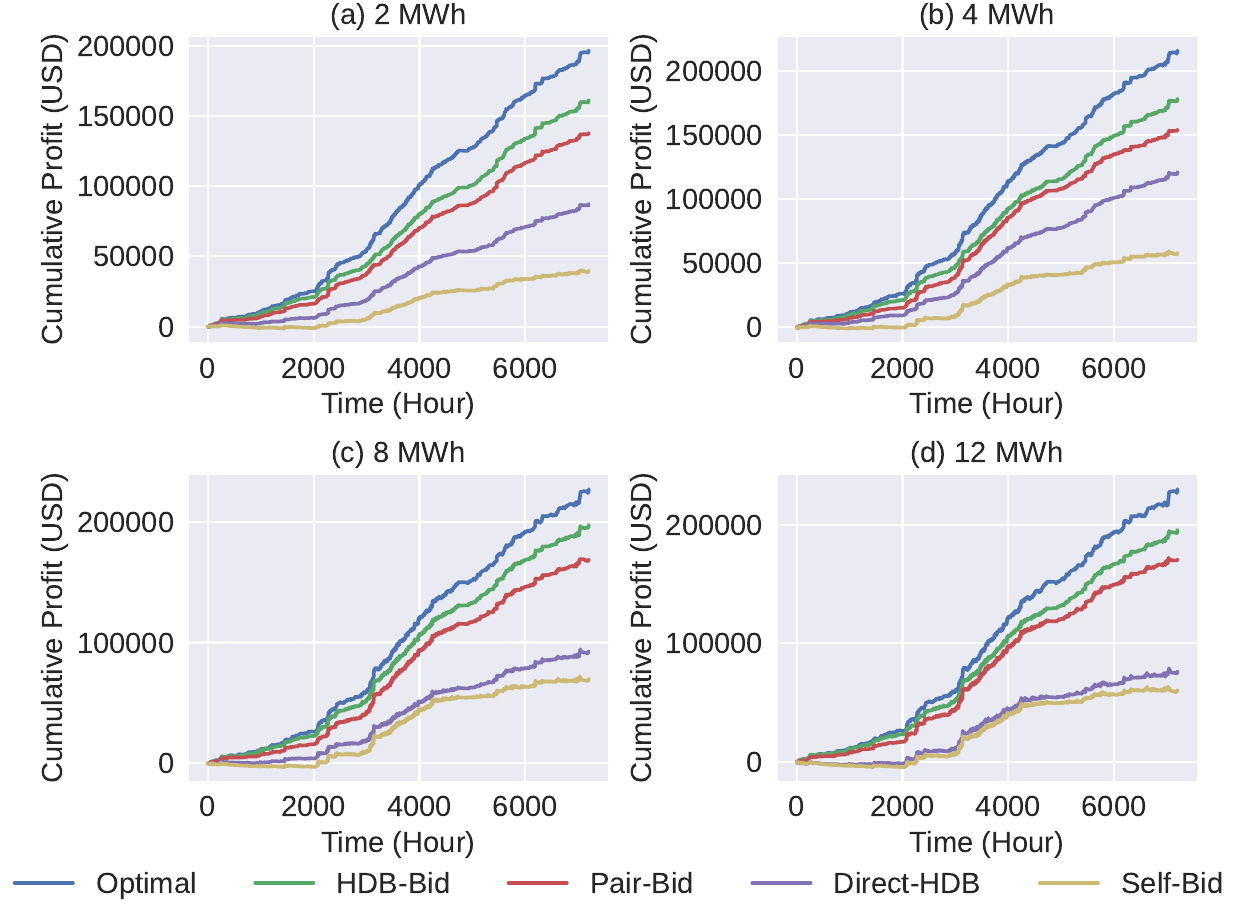}
    \caption{The cumulative test reward of different bidding methods at the PJM-RTO price node, shown at different energy storage capacities}
    \label{fig:culmulative_reward}
\end{figure}

% The improvement of optimal profit from 2MWh to 12MWH is not significant. The optimal profit of 2MWh is 196695.76USD while the optimal profit of 12MWh is 229876.92USD. Only a 16.87\% percent increase with 5 times large energy storage capability. Therefore, it is more economically viable to opt for short-term ESS in RT energy markets.

% HDB 把Pair Bid距离最优的收益缩短了接近一半
The HDB-Bid fills the gap from Pair-Bid (previously the most flexible bid) to the optimal bidding income by an average of 43.90\%.
The gap to the optimal profit for Pair-Bid is  28.31\% on average, and for HDB-Bid is 15.88\%. 
The gap is filled by 43.90\%, which is a significant boost in terms of bidding profits.
\color{black}

\color{black}
\subsubsection{Bid Generation Efficiency}

The efficiency of the HDB generation algorithm (\cref{alg:bid_discretization}) is evaluated in this subsection.

Previous RL-based bidding methods generate market LDB by running a neural network forward pass. The neural network's output is the market LDB, such as the one-value power bids\cite{wei_self-dispatch_2022}.
Comparatively, the proposed HDB bidding method generates each market HDB with the HDB generation algorithm (\cref{alg:bid_discretization}).
For each HDB in \cref{alg:bid_discretization}, the neural network is run $M=(\lambda_{\max}-\lambda_{\min})/\delta\lambda$ times to sample the supply curve. 
In our case studies, the sample resolution $M=512$.
Then, an HDB extraction process is performed to extract the HDBs.

To evaluate the efficiency of the proposed HDB generation algorithm, we benchmark the runtime of different bid generation methods. The computation time to generate one market bid is computed by averaging over 1000 runs. In \cref{alg:bid_discretization}, we also investigate neural network parallelization of batch\_size 1 to $M$ in the supply curve sampling process.
The run time comparison is shown in \cref{tab:time_comp}

% \usepackage{graphicx}
\begin{table}[h]
\centering

\caption{Generation time of generating a market bid}
\label{tab:time_comp}

\resizebox{\columnwidth}{!}{%
\begin{tabular}{c|c|c|c|c|c|c}
\hline
Method & LDB & \begin{tabular}[c]{@{}c@{}}HDB \\ batch\_size=512\end{tabular} & \begin{tabular}[c]{@{}c@{}}HDB \\ batch\_size=16\end{tabular} & \begin{tabular}[c]{@{}c@{}}HDB \\ batch\_size=4\end{tabular} & \begin{tabular}[c]{@{}c@{}}HDB \\ batch\_size=1\end{tabular} & HDB extraction \\ \hline
Time   & 0.570ms & 4.081ms & 6.950ms & 13.967ms & 42.411ms & 3.045ms \\ \hline
\end{tabular}%
}

\end{table}

The  HDB generation time consists of two parts. The supply curve sampling part and the HDB extraction part.
The supply curve sampling process scales linearly w.r.t. the forward pass times of the neural network, ranging from $1ms$ to $40ms$.
The HDB extraction part has a constant time of about $3.045 ms$.

The results show that parallelization is important for speeding up the HDB generation process.
Parallelization of neural network forward can be easily achieved on a single GPU by setting a proper batch size.
As a result, even though the proposed approach will consume more computation resources, it still has a similar run time when parallelization is possible.

In summary, the HDB generation has a running time of milliseconds on the mentioned hardware. By neural network forward pass parallelization, the process can reach a running speed of $4.081ms$, which is efficient for hourly power market bidding.

% 收益可视化
\subsection{Bidding Visualization}

% 简介可视化
In this subsection, we visualize the HDB bidder's bidding history for a more intuitive understanding of its high-dimensional bid decisions.

\subsubsection{Bidding process visualization}
\cref{fig:income_history} provides a demonstration of a 2-day bidding result on the PJM-RTO price node with different energy storage capacities.

\begin{figure*}[h]
    \centering
    \includegraphics[width=1\linewidth]{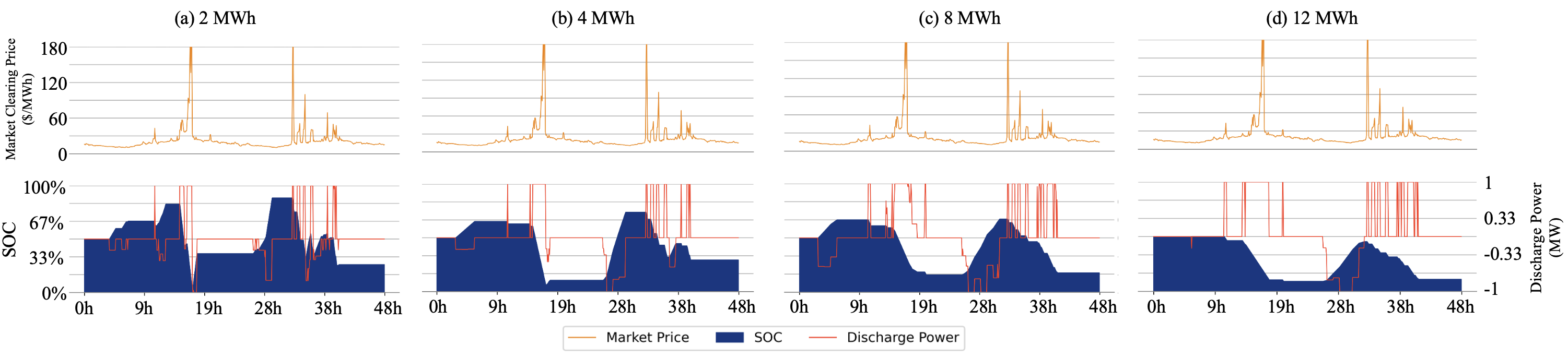}
    \caption{A sample bidding history of a 1, 2, 4, 8MWH ESS in the PJM-RTO price node. Yellow lines are market clearing prices, red lines are discharge powers of ESS, black areas are the SoC of the ESS.}
    \label{fig:income_history}
\end{figure*}

% 详细介绍图片
The upper subfigures show the RT energy market prices. The lower subfigures show the SoC histories in black areas and show the market clearing power in red lines. Positive power means discharging, and negative power means charging.
We can observe that the RT market price is highly uncertain, and the price peaks are difficult to predict\cite{zhang_predicting_2022,yang_qcae_2022}.

% 简单介绍 discharging behavior
Overall, the bidder is able to bid strategically.
From the perspective of charging (negative power values), the bidder is able to capture the low prices during low-price hours at night. It charges the SoC to proper levels before high-price times. 
% All bidders are able to properly utilize the lowest price hours to charge the ESS  during the night at the valley hours, and the charging behaviors can be observed to be near-optimal from an ex-ante perspective. The final charged SoC of different energy storage capacities is different. Smaller ESS capacity charges the SoC to a higher level.
% 简单介绍 charging behavior
From the perspective of discharging (positive power values), the bidders are able to precisely capture the high-price peaks and allocate the stored energy among them. 
\color{black}During the high-price times, the 2MWh ESS (\cref{fig:income_history}.a) discharges at the high-price peaks and charges at normal prices to arbitrage. Comparatively, the 12MWh ESS (\cref{fig:income_history}.d) only discharges during the high-price peaks because it stores enough energy in the low-price hours. Both behaviors are plausible from an ex-ante perspective.\color{black}

% 汇总不同soc行为的区别（提到soc越界的问题）
By comparing the power routes and SoC routes of different ESS, we can see that the proposed bidder adapts to the ESS capacity with proper strategies to charge at low market prices and discharge at high market prices.
\color{black}

\subsubsection{HDB visualization}
The generated HDBs corresponding to \cref{fig:income_history} is demonstrated in \cref{fig:bid-demo}. The HDBs for different energy storage capacities are shown in 3D plots. The vertical axis is the discharge power in $MW$, and the horizontal axes are the bid prices and bid time.

\begin{figure}[h]
    \centering
    \includegraphics[width=1\linewidth]{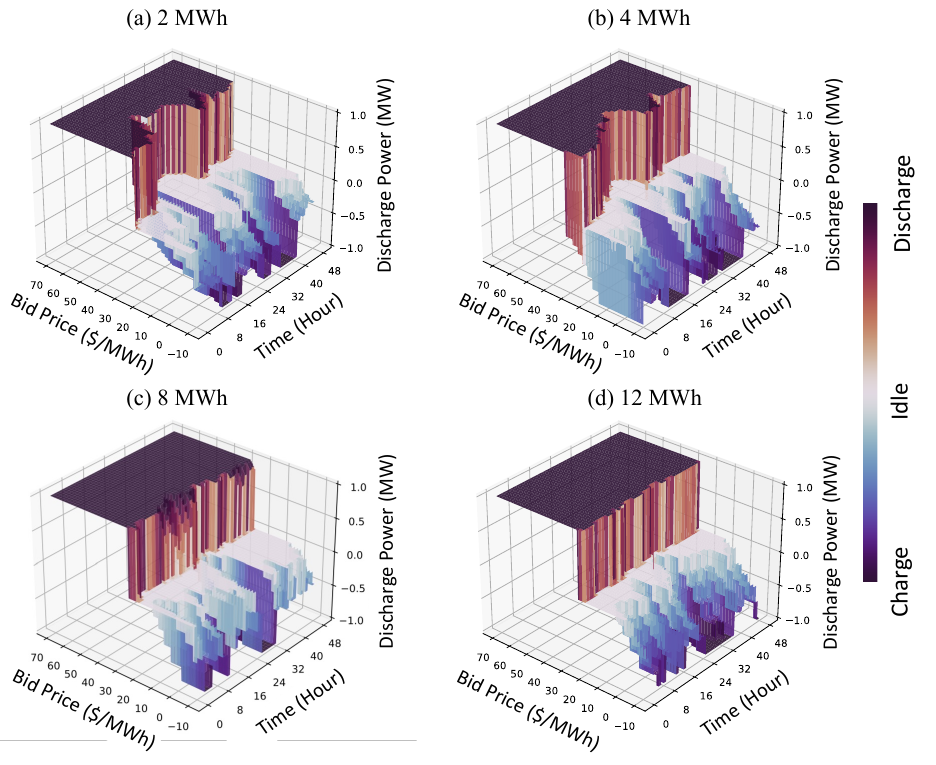}
    \caption{Demonstration of the historical HDBs corresponding to \cref{fig:income_history}. The vertical axis is the bidding power. The horizontal axis is the bid price and the time index. One bid corresponds to one hour.}
    \label{fig:bid-demo}
\end{figure}

From the perspective of discharging (upper red regions), the bidders usually have a clear price threshold for discharging power. The bidders will choose to discharge above a specific price threshold and choose to be undispatched under that threshold. The price threshold is different for ESS with different energy storage capacities. In general, a low-capacity ESS has a more sensitive price threshold that changes with the bidding time. while high-capacity ESS has a more stable price threshold that rarely changes throughout the day. This can be explained by the fact that the change in SoC percentage is deeper for a low-capacity ESS. Therefore, low-capacity ESS changes the price threshold according to the change in energy levels.

From the perspective of staying idle (middle white regions), the bidders are able to withhold energy strategically. The white regions in \cref{fig:bid-demo} denote the time and price that the agents choose to be undispatched. ESS with different energy storage capacities will adopt different idle strategies. From the power dispatch results in \cref{fig:income_history}, we can see that the ESS is undispatched for most of the bidding time, rendering the importance of such a strategy.

From the perspective of charging (lower black regions), the bidders utilize the high-dimensionality of HDBs for price-responsive discharge. Bidders will charge different energy for different market prices. The low-capacity bidders are less price-sensitive, as they will conduct charging even if the market price is higher than 20USD/MWh. The high-capacity bidders are more price-sensitive, as they only conduct discharge for prices below 20USD/MWh. This can be explained by their different capability to capture the low prices for charging. Because high-capacity ESS is able to charge large amounts of energy at low prices, they don't need to buy energy when the market price is high. However, low-capacity ESS cannot store much energy during low prices, so they have to buy energy during specific high-price scenarios.

% In the two scenarios, the HDBs show different patterns. For the discharging bids (red region), the left figure has a clear price threshold from zero discharge to full discharge, while in the right figure, there is a transit price range from zero discharge to partial discharge to full discharge. We explain this phenomenon as the two scenarios have different opportunity costs for discharge. In the left figure (low price), because there is little opportunity to discharge, the opportunity cost is low. Therefore, as long as the market price is higher than the marginal generation cost, the best decision is full discharge. In the right figure, the opportunity cost is high because there are a lot of discharge opportunities. Therefore, it is better to choose partial discharge for prices that are not very high.

% For the charging bids (black region), the two scenarios have opposite behaviors. The right figure has a clear price threshold from zero charge to partial charge to full charge, while the left figure has a transit price region. This can also be explained by the opportunity costs of the discharging direction. Because the right figure needs more power for discharge, the opportunity cost for charging is low, and its charging action is more direct. On the left figure, because there are a lot of opportunities to charge, the opportunity cost for charging is higher, and the agent chooses partial charging for prices that are not very low.

From the above experiments, we can see that HDBs can be effectively generated by the proposed bidder, the HDB’s flexibility can be strategically utilized, and the generated HDBs have good interpretability.
In the following subsection, we will further discuss the relations between the NNSF and the HDBs.

\subsection{The Influence of Approximating NNSFs with HDBs}

In \cref{sec:HDBgen}, we propose an HDB generation algorithm to generate HDBs from NNSFs.
The algorithm includes a supply curve sampling process and an HDB extraction process.
The supply curve sampling process samples the input-output relationship of the NNSF as a supply curve.
It is accurate because the sampling resolution can be arbitrarily increased to meet the precision needs.

However, the HDB extraction process will inevitably cause errors because the extracted 2N-dimensional HDB will lose information from the supply curve. 
The HDB extraction process includes three steps: the monotonize, discretize, and output bid process. (\cref{fig:bid_fact})
The monotonize step will modify the supply curve if it is not monotonic.
The discretize steps will cause approximation errors in extracting 2N-dimensional HDBs from supply curves.

In this subsection, we will further discuss the degree of these two errors and how such approximation errors influence the bidding results.
Overall, we found that the monotonize step is beneficial to bidding, and the discretize step is minor detrimental.

\subsubsection{Comparing the Performance of NNSF and HDB}

The bidding profit of NNSF and HDB are compared in \cref{tab:hdb-nnsf}. They apply the trained NNSF to the training bidding process of \cref{fig:learning_framework} (NNSF), and the HDB generation process of  \cref{fig:generation_framework} (HDB). Because the training bidding process directly dispatches the ESS with NNSF, it reflects the direct bidding result of NNSF without HDB generation.

\begin{table*}[h]

\caption{The Captured profit ratio of HDB (approximated from NNSF) and NNSF}
\label{tab:hdb-nnsf}

\centering
\begin{tabular}{l|ll|ll|ll|ll|ll}
\hline
         & \multicolumn{2}{c|}{PJM-RTO} & \multicolumn{2}{c|}{DOM} & \multicolumn{2}{c|}{EKPC} & \multicolumn{2}{c|}{PSEG} & \multicolumn{2}{c}{COMED} \\ \hline
Capacity & NNSF          & HDB          & NNSF        & HDB        & NNSF        & HDB         & NNSF        & HDB         & NNSF        & HDB         \\
2MWh     & 77.35\%       & 81.94\%      & 76.20\%     & 83.51\%    & 74.96\%     & 81.35\%     & 70.34\%     & 77.49\%     & 72.19\%     & 76.73\%     \\
4MWh     & 78.01\%       & 82.46\%      & 78.28\%     & 86.16\%    & 80.94\%     & 86.07\%     & 70.04\%     & 79.31\%     & 76.53\%     & 81.21\%     \\
8MWh     & 81.29\%       & 86.99\%      & 79.24\%     & 87.11\%    & 83.07\%     & 88.41\%     & 68.32\%     & 75.26\%     & 78.58\%     & 82.71\%     \\
12MWh    & 80.21\%       & 85.08\%      & 79.30\%     & 87.44\%    & 80.31\%     & 86.35\%     & 62.01\%     & 70.84\%     & 77.56\%     & 82.12\%     \\ \hline
\end{tabular}

\end{table*}

Surprisingly, the HDB achieves higher bidding profit than NNSF.
Intuitively, the HDB generation algorithm loses information from the original NNSF. This would cause a performance drop due to the loss of information.
However, the test results show that the performance increased by  4.13$\sim$9.27\%, which is a substantial improvement.

In the following subsections, we find out the reason for such improvement by examining the influence of the monotonize and discretize steps.

% 单调性
\subsubsection{The Influence of the Monotonize Step of the HDB Generation Algorithm}

In this subsection, we will demonstrate that the Monotonize step in \cref{alg:bid_discretization} is overall beneficial for the bidding results. We will discuss its influence on both the normal price ranges and the extreme price ranges.

First, on the normal price ranges (-50USD, 200USD), the learned NNSF $\pi_\theta(\hat{s}_t,\lambda_t)$ is monotonic in general. Therefore, the monotonize step will not cause much approximation error for normal price ranges.
We will use two metrics to demonstrate the NNSF's monotonicity.

The first metric is the number of monotonic supply curves, which corresponds to the percentage of supply curves that are completely monotonic and do not need to be monotonized.
The second metric is the percentage of supply curve points that need to be changed to achieve total monotonicity. This describes how much the monotonicity assumption is violated from a point-based view and describes how much deviation is introduced in monotonizing the supply curves.

On the first metric, the percentage of completely monotonic supply curves is 98.24\%. This shows that only 1.76\% of supply curves are affected by the monotonize step, and most supply curves are monotonic. On the second metric, the percentage of monotonic supply curve segments is 99.80\%. This shows that the monotonize step will only change 0.2\% of the total supply curve segments. 

The change of 0.2\% change in supply curve segments is relatively negligible from the perspective of the whole NNSF. As a result, the error of the monotonize step (step b in \cref{alg:bid_discretization}) for normal price ranges is small.

Second, on the extreme price ranges (prices beyond the normal price range), the learned NNSF is generally not monotonic.
The percentage of completely monotonic supply curves between price 200USD and 500USD is 40.29\%, and the percentage of monotonic bidding segments is 62.01\%.

The reason for such non-monotonicity is that extreme cases are difficult to train in machine learning. Because extreme prices are rare in the training dataset, they do not have enough training data, so these cases are not fitted enough in the training process. 
Also, because the extreme cases are numerically ill-conditioned data that deviates from the normal price distribution, they are hard to fit by neural networks.

Therefore, the monotonize step will change the actions at extreme price ranges. However,  the change is beneficial for bidding.
The monotonize step will overwrite the action of the extreme price ranges with the action of the normal price ranges, which involves full discharge power for extremely high prices and full charge power for low prices.
It helps the bidding agent to achieve plausible actions for extreme prices.

As a result, the monotonize step is beneficial overall for bidding performance. The average captured profit percentage of the monotonized supply curve is 82.59\%, which is higher than the HDB-Bid (82.43\%) and the NNSF (76.24\%).
In the following context, we will use N=$\infty$ to denote the monotonized supply curve, which is a discretized supply curve with infinite resolution.

% 离散型（离散误差有多少）
\subsubsection{The Influence of the Discretize Step of the HDB Generation Algorithm }

In this subsection, we will show that the discretize step in \cref{alg:bid_discretization} will negatively impact the bidding performance, but the influence is minor because the HDB is high-dimensional and flexible.
We will compare the performances of the monotonized supply curve  with different dimensions of HDBs. Then, we will measure the discretization errors from the monotonized supply curve to HDB.

\begin{figure}[h]
    \centering
    \includegraphics[width=1\linewidth]{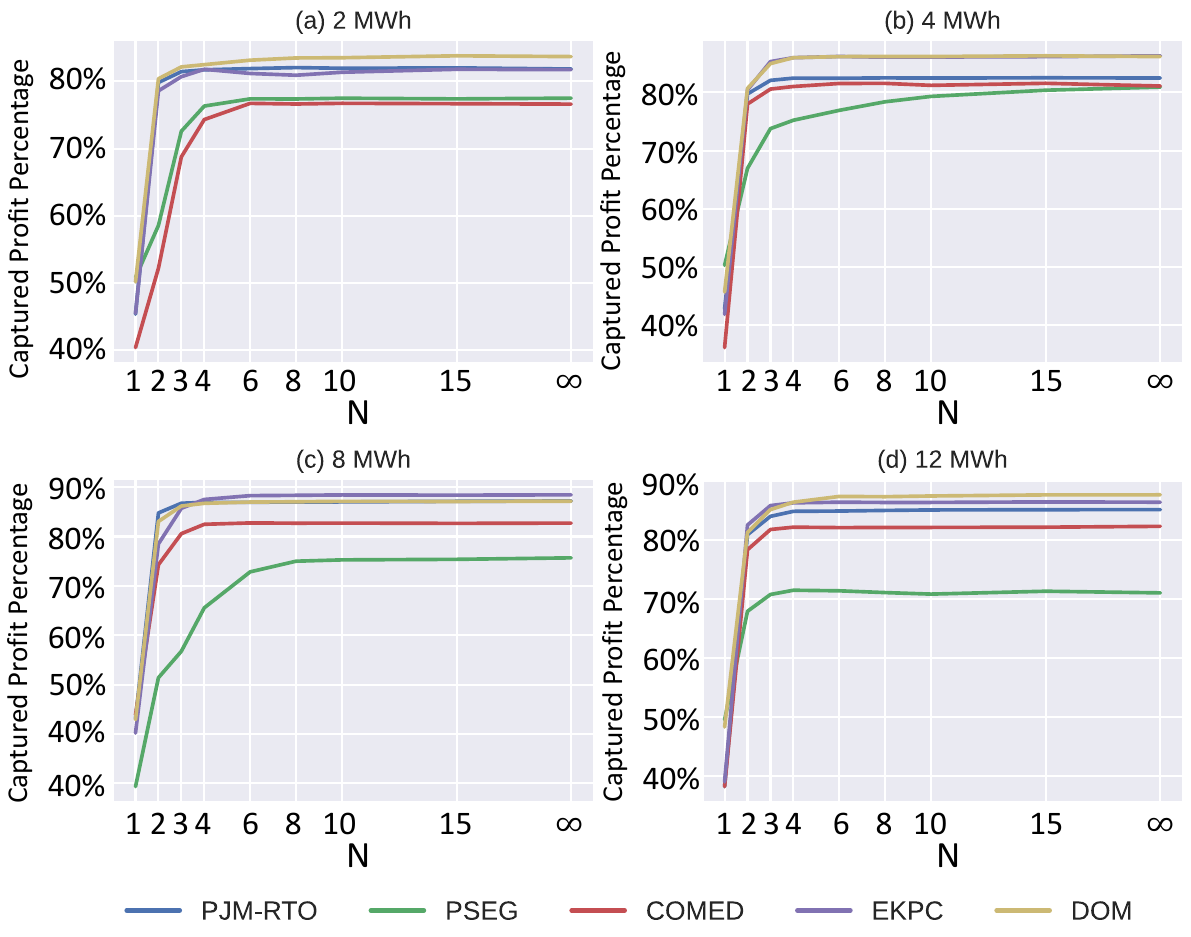}
    \caption{The captured profit ratio of HDBs with different dimensions. The horizontal axis is N, which is the degree of freedom of the bidding pairs.}
    \label{fig:n_bid_compare}
\end{figure}

The bidding performance of HDBs with different dimensions is shown in \cref{fig:n_bid_compare}. The bidding performance is shown in the captured profit ratios and grouped by SoC values.

In general, a higher HDB dimension (a higher N) achieves a better bidding performance. 
The N=1 case has the lowest captured profit because its flexibility is the lowest. The average capture profit percentage of N=1 across all experiments is 43.10\%.
From N=2 to N=6, the captured profit percentage of HDBs gradually increases. The average captured profit percentage is  74.91\%(N=2), 79.78\%(N=3),  81.37\%(N=4) and 82.16\%(N=6). 
For N$\ge$6, the profit is mostly saturated, and the performance increases slowly.  The average captured profit percentage is 82.34\%(N=8),  82.43\%(N=10),  82.58\%(N=15) .
Finally, the monotonic supply curve (N=$\infty$) has a captured profit ratio of 82.59\%.

N=10 is the most common HDB format, and its performance is close to N=$\infty$(no discretization). The gap from N=10 to N=$\infty$ is 0.16\% profit percentage,  which means the common HDBs can capture sufficient profits for the bidding.

N=2 corresponds to the Pair-Bid bidding format\cite{tao_deep_2022}. Its performance gap to N=10 is 7.52\%, and its performance gap to N=$\infty$ is 7.68\%. This shows that the Pair-Bid format could lose a considerable amount of profit in real-world bidding.

Additionally, we quantify the errors of the discretize step by comparing the market clearing power of HDB-Bid (N=10) and the monotonized supply curve (N=$\infty$).
We measure the  mean absolute error (MAE) of market clearing power from monotonic supply curves to HDBs.
The MAE is 0.0103 (with full power as $\pm$1), which is 1.03\% of the maximum power.

Overall, the discretize step will introduce approximation errors to the bidding process, but the approximation error and the performance drop of most market's HDBs is minor for real-world bidding.
\color{black}

\subsection{Discussions and Further Applications}

From the above experiments, we can see that using NNSF for training and generating HDBs for real-world bidding can produce high-performance bidding results. 
Though the framework includes various approximations, including the monotonize and discretize steps in \cref{sec:bid-generation}, and the market clearing power approximation in \cref{sec:learning-framework}, the bidding results are satisfactory.

This phenomenon relates to the problem's characteristics. 
On the one hand, the discretize step and the market clearing power approximation are accurate because the HDB is a high-dimensional bid, and it can represent the NNSF with high precision.
On the other hand, the monotonize step is accurate because the bidder is profit-seeking, and it will tend to bid more power for a higher market clearing price in normal price ranges, and it is originally monotonic in most cases.
As a result, the proposed training process is accurate for training NNSFs and generating HDBs.

The proposed framework provides a new way for RL-based HDB bidding: we can learn an NNSF first and use HDBs to approximate it. So that HDBs can be learned and generated effectively.

\color{black}
Regarding the RL problem's structure, the proposed framework transforms the market price $\lambda_t$ from an unknown input to a known input of the neural network agent.
Which makes the bidding problem a simpler task for RL algorithms to learn.
\color{black}

In the future, the proposed NNSF process can be applied to other HDB bidding scenarios, such as the Day-Ahead energy market, the regulation market, the reserve market, etc.

\section{Conclusion}\label{sec:conclusion}

This paper proposes a bidding framework that effectively utilizes HDBs for the first time in RL-based power market bidding methods.
Though the HDB is the most common market bidding format in the form of N price-power pairs, past RL-based methods have failed to fully utilize the HDB for power market bidding due to its high dimensionality.
The loss of flexibility in current RL bidding methods could greatly limit bidding profits and make it difficult to tackle the rising uncertainties brought by renewable energy generations.
To tackle the above challenges, we propose a framework that is suitable for RL-based methods with HDB bidding.
First, we employ a special kind of neural network called NNSF to construct a strategic mapping from market clearing price to bidding power.
Second, we propose a generation framework to extract HDBs from the input-output relation of NNSF.
Then, we propose an approximation of the generation framework and compose a new training framework, which is compatible with most state-of-the-art RL algorithms.
Finally, a PPO-based RL algorithm is employed to train an HDB bidding policy for an ESS in the RT energy market.
Experiment results show that the proposed algorithm is able to efficiently leverage the bid flexibility of HDBs and generate strategic HDB bidding pairs.
The proposed algorithm can improve the bidding performance based on state-of-the-art RL methods by an average of 15.40\% and reaches 70.84\%$\sim$88.41\% optimal market profit, which is the highest reported profit ratio in the literature captured by RL-based methods for RT energy market bidding.
The future work includes empowering a larger number of RL-based power market bidding methods with the flexibility of HDBs using the proposed bidding framework.

\bibliographystyle{elsarticle-num} 
\bibliography{references.bib}

\end{document}